\documentclass{article} 
\usepackage{iclr2026_conference,times}

\usepackage{amsmath,amsfonts,bm}

\def\eqref#1{equation~\ref{#1}}
\def\Eqref#1{Equation~\ref{#1}}
\def\plaineqref#1{(\ref{#1})}

\def\1{\bm{1}}

\DeclareMathAlphabet{\mathsfit}{\encodingdefault}{\sfdefault}{m}{sl}
\SetMathAlphabet{\mathsfit}{bold}{\encodingdefault}{\sfdefault}{bx}{n}

\newcommand{\E}{\mathbb{E}}
\newcommand{\Ell}{\mathcal{L}}
\newcommand{\R}{\mathbb{R}}

\usepackage[hidelinks]{hyperref}
\usepackage{url}
\usepackage{amsmath}
\usepackage{amssymb}
\usepackage{wrapfig}
\usepackage{subfigure}
\usepackage{graphicx}
\usepackage{subcaption}
\usepackage{soul}
\usepackage{units}

\title{Sensitivity Analysis for Diffusion Models}

\author{Christopher Scarvelis \\
MIT CSAIL \\
Cambridge, MA \\
\texttt{scarv@mit.edu} \\
\And
Justin Solomon \\
MIT CSAIL \\
Cambridge, MA \\
\texttt{jsolomon@mit.edu} \\
}

\newtheorem{theorem}{Theorem}[section]

\newtheorem{lemma}[theorem]{Lemma}

\iclrfinalcopy 
\begin{document}

\maketitle

\begin{abstract}

Training a diffusion model approximates a map from a data distribution $\rho$ to the optimal score function $s_t$ for that distribution. Can we differentiate this map? If we could, then we could predict how the score, and ultimately the model's samples, would change under small perturbations to the training set before committing to costly retraining. We give a closed-form procedure for computing this map's directional derivatives, relying only on black-box access to a pre-trained score model and its derivatives with respect to its inputs. We extend this result to estimate the sensitivity of a diffusion model's samples to additive perturbations of its target measure, with runtime comparable to sampling from a diffusion model and computing log-likelihoods along the sample path. Our method is robust to numerical and approximation error, and the resulting sensitivities correlate with changes in an image diffusion model’s samples after retraining and fine-tuning.

\end{abstract}

\section{Introduction}\label{sec:intro}


Diffusion models form a powerful class of generative models that allow users to generate images of nearly any subject in nearly any style in just a few keystrokes. However, this flexibility also allows diffusion models to engage in legally fraught behavior, such as generating images that mimic an artist's style. This has put diffusion models at the center of recent litigation\footnote{\textit{Andersen v. Stability AI Ltd.}, U.S. District Court, Northern District of California (2024).} alleging that they facilitate copyright and trademark infringement. Understanding and mitigating the causes of this behavior have therefore become pressing challenges as businesses seek to integrate diffusion models into their consumer offerings.

Diffusion models generate images by iteratively transforming Gaussian noise using the \emph{score function} of a distribution over noisy images, which is learned in practice from a large set of training images. Since the learned score is, in principle, determined by the training images, a natural strategy for understanding a diffusion model's behavior is to study how it depends on the training data. A sensible framework for this task should be able to answer questions such as: ``What would the score function be if a sample were added or removed from the training set?'' and ``What would a generated image have looked like if a sample were added or removed from the training set?''

This work introduces a principled framework for answering such questions about diffusion models in the perturbative regime, where one considers infinitesimal changes in a model's training distribution. 
Because a diffusion model's output depends on the score function, which is itself determined by the training distribution, the core of our framework is a tractable closed-form expression for the directional derivatives of a score function with respect to the training distribution. 
This \emph{sensitivity analysis} measures how the score function changes as a probability measure is up- or down-weighted in the training distribution; when this measure is a Dirac point mass, we obtain an exact expression for the \emph{influence function} of the score. Crucially, our sensitivity analysis requires only black-box access to a pre-trained score function, and it does not require any knowledge of its training data or training procedure.

Using the adjoint method, we extend our sensitivity analysis for score functions to obtain schemes for computing the sensitivity of a diffusion model's samples to perturbations in the training data. These enable us to predict how a generated image would change if a collection of samples were added or removed from the training set. We demonstrate our method's robustness to a variety of sources of numerical error and show that its predictions are correlated with changes in a diffusion model's samples after retraining and after fine-tuning. 

\section{Related work}\label{sec:related-work}

\paragraph{Influence functions.}

Influence functions \citep{hampel74} linearly approximate the change in a statistical estimator in response to infinitesimally upweighting a single training sample. \citet{koh2017understanding} introduced influence functions to deep learning as a method for estimating the change in a neural network's parameters in response to perturbing its training set. Influence functions for generic optima of a training loss require the user to compute a costly inverse Hessian-vector product. Previous work \citep{guo-etal-2021-fastif, schioppa2022scaling} responds to this challenge by developing efficient approximations to this operation. In addition to this difficulty, influence functions assume that the learned model parameters minimize a strictly convex loss. This assumption is violated for neural networks, and
\citet{Basu2020InfluenceFI} find that in practice, influence functions for deep learning are brittle to network hyperparameters. \citet{kwon2024datainf, mlodozeniec2025influence} introduce influence approximations that are specially adapted to generative models, including diffusion models, but these works follow \citet{koh2017understanding} in estimating the influence of training samples on the learned network weights. In contrast, our sensitivity analysis uses the structure of diffusion models to directly compute the influence of training samples on the value of the score function and on model samples.

\paragraph{Data attribution for diffusion models.} 

An emerging literature develops \emph{data attribution} methods for diffusion models, which seek to estimate the impact of training samples on model outputs. \citet{georgiev2023journey} use TRAK \citep{park2023trak}, a gradient-based data attribution method developed primarily for supervised learning, to compute a per-example attribution score for a diffusion model's training data. This score estimates the change in the model's training loss induced by adding a particular sample to the training set. Following \citet{park2023trak}, they measure their method's effectiveness using the \emph{linear datamodeling score}, which measures the rank correlation between their attribution score and actual training loss values attained by retrained models. \citet{zheng2024intriguing} observe that one can improve upon the method from \citet{georgiev2023journey} by computing their attribution scores using the gradients of the ``wrong'' model output function. \citet{mlodozeniec2025influence} introduce an efficient approximation to the denoising loss Hessian and use it to estimate influence functions for attributing several proxies for model log-probabilities. \citet{lin2025diffusion} propose attributing the KL divergence between the model distribution before and after deleting a training sample. \citet{li2025learning} perform gradient-based data attribution using learnable weights for gradients with respect to different parameter groups. Whereas these methods estimate the impact of training samples on scalar quantities such as the training loss or proxies for log-probabilities, our sensitivity analysis estimates the effect of perturbations in the target distribution on the values of the score function and on model samples.

\section{Method}\label{sec:method}

In this section, we begin by observing that a diffusion model defines a map from its training distribution $\rho$ to a score function $s_t$ and then show how to tractably compute its directional derivatives. We then use this result to estimate how a diffusion model's samples change in response to small perturbations of its training distribution.

\subsection{Preliminaries}\label{sec:prelims}

\emph{Diffusion models} sample from a target distribution $\rho$ by drawing samples from a Gaussian base distribution $\mathcal{N}(0,I)$ and flowing them through a possibly noisy velocity field $v_t$ from $t=t_0$ to $t=t_1$. This yields a curve of probability distributions $\{\rho_t : t \in [t_0,t_1] \}$ for which $\rho_t$ is the marginal distribution of the random variable $Z_t := \alpha_tX_1 + \sigma_t \epsilon$. Here, $X_1 \sim \rho$,  $\epsilon \sim \mathcal{N}(0,I)$, and $\alpha_t$ and $\sigma_t$ are scale and noise schedules, respectively. 
These schedules are chosen so that at $t=t_0$, the samples are distributed according to a Gaussian distribution, and at $t=t_1$, the samples are distributed according to the target distribution $\rho$.

A diffusion model's velocity field $v_t$ depends on $\rho$ through the \emph{score function} $s_t(z) := \nabla \log \rho_t(z)$ of $\rho_t$, which one learns in practice by solving a \emph{score-matching} problem \citep{hyvarinen2005estimation}. If one does not impose any restrictions on the hypothesis class, the optimal solution to this problem is in fact available in closed form \citep{miyasawa1961}:
\begin{equation}\label{eq:score-function}
    s_t(z) \!=\! \frac{1}{\sigma_t^2}\!\left(\int \!w_t(z,x)\, \alpha_t x \, \textrm{d}\rho(x)\!-\!z\! \right)\!, 
    \text{where }
    w_t(z,x) \!:=\! \frac{\exp\left(\!-\frac{1}{2\sigma_t^2}\|z - \alpha_t x\|_2^2\right)}{\int\!\exp\left(\!-\frac{1}{2\sigma_t^2}\|z - \alpha_t x\|_2^2\right)\textrm{d}\rho(x)}.
\end{equation}

\begin{wrapfigure}[11]{r}{0.6\textwidth} 
  \centering\vspace{-.18in}
  \begin{minipage}{0.2\textwidth}
    \centering
    \includegraphics[width=\linewidth]{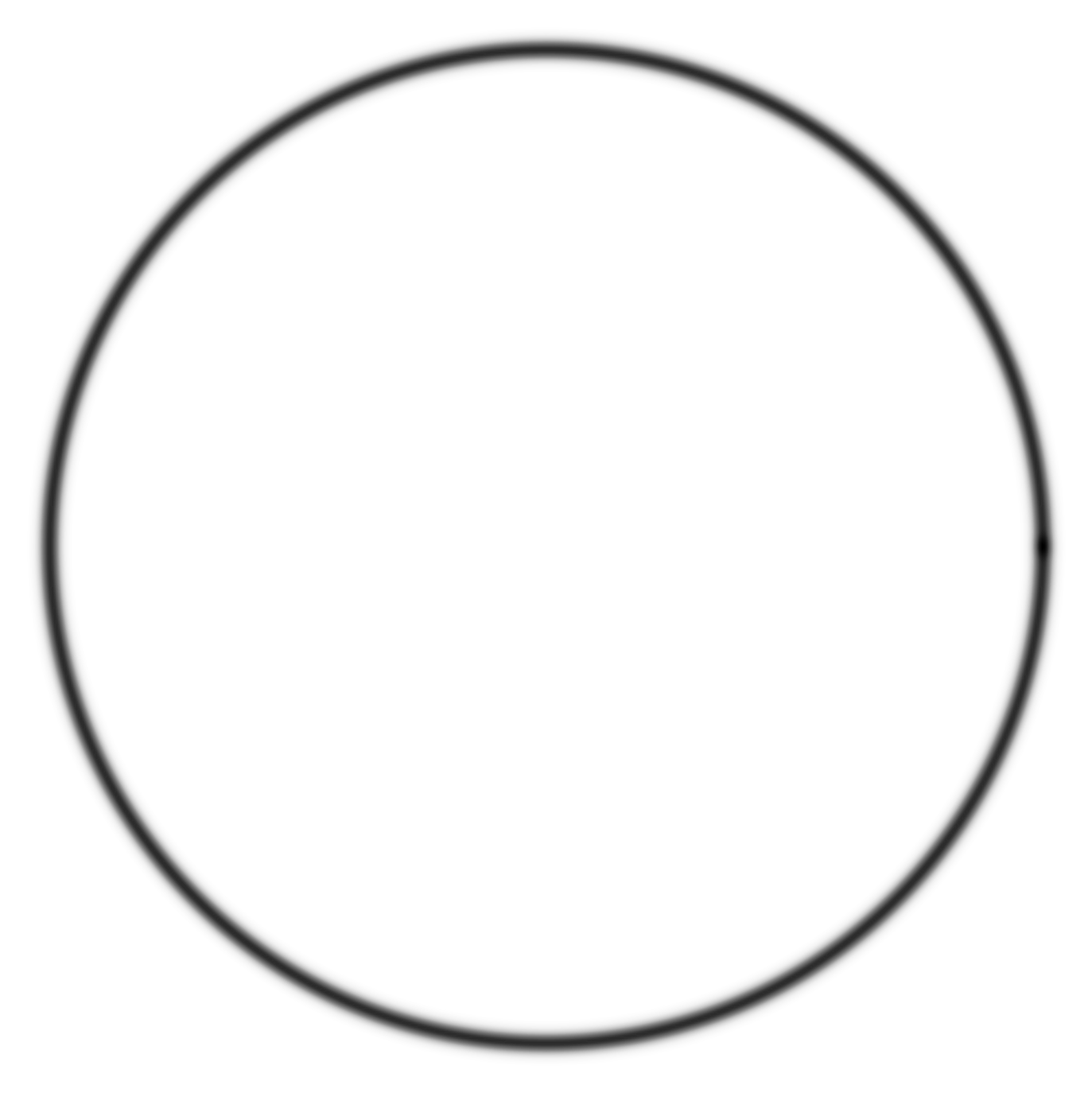}\label{fig:circle-density}
  \end{minipage}
  \quad 
  \raisebox{0.4\height}{\textbf{\boldmath$\Longrightarrow$}}
  \quad
  \begin{minipage}{0.2\textwidth}
    \centering
    \includegraphics[width=\linewidth]{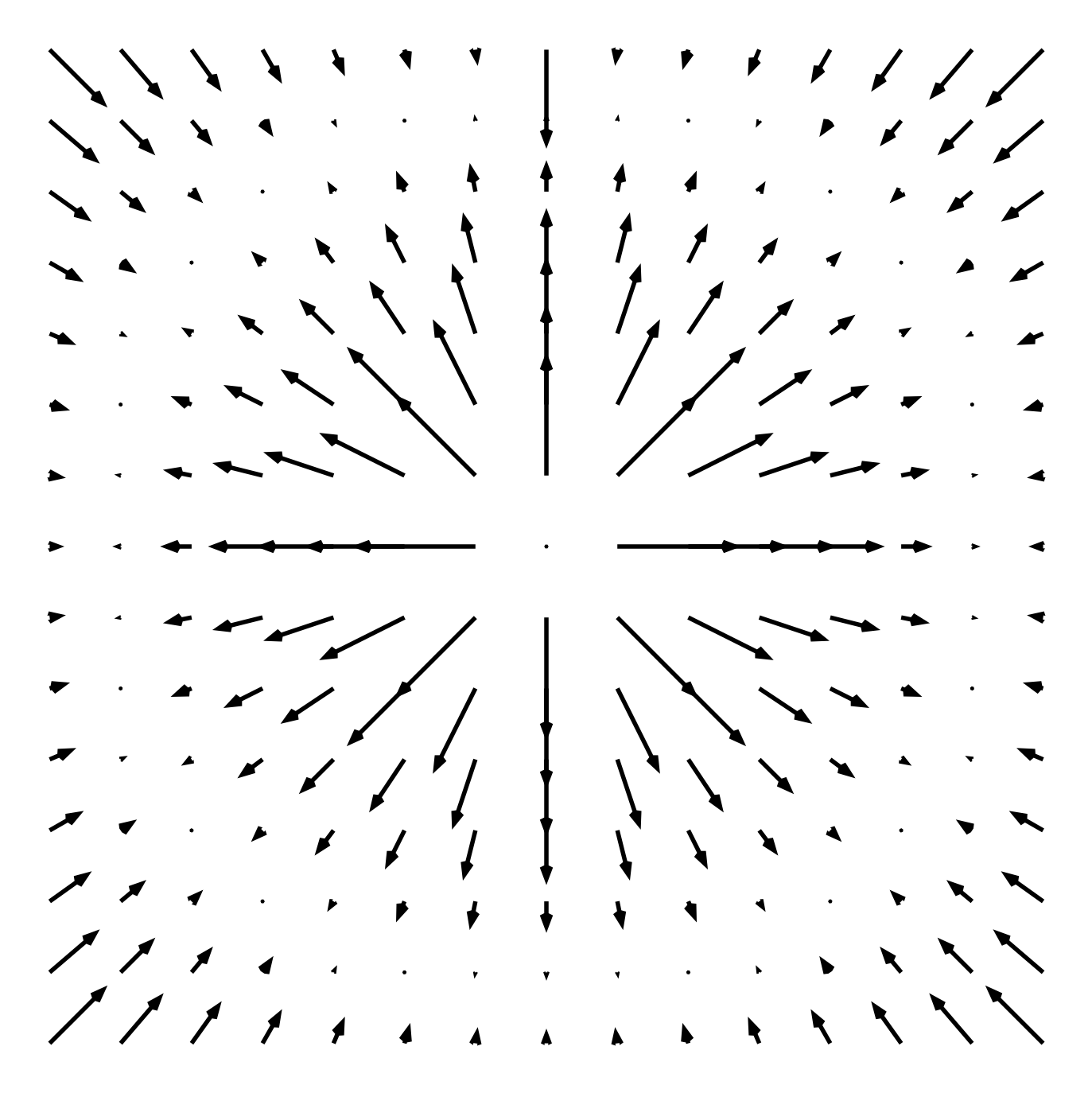}\label{fig:second}
  \end{minipage}
  \caption{Solving the score-matching problem maps a measure $\rho$ (left) to a vector-valued function $s_t$ (right). We will compute this map's directional derivatives in Section \ref{sec:score-sensitivity}.}
  \vspace{-12pt}
  \label{fig:measure-to-score}
\end{wrapfigure}

This is a vector field pointing from $z$ toward a distance-weighted average of rescaled samples $\alpha_t x$ from the target distribution $\rho$. Solving the score-matching problem therefore maps a \emph{measure} $\rho$ to a \emph{function} $s_t$, which is fully determined by $\rho$ and the scale and noise schedules.

We would like to estimate how the outputs of a diffusion model would change in response to perturbations of the training data. These outputs depend on the training data only through the velocity field $v_t$ and, in turn, through the score function $s_t$. We will therefore begin by introducing a tractable closed-form expression for the directional derivatives of the map from $\rho$ to $s_t$, which will describe how $s_t$ responds to additive perturbations of the target distribution $\rho$.

\subsection{Sensitivity analysis for score functions}\label{sec:score-sensitivity}

\Eqref{eq:score-function} expresses the score function $s_t$ of a diffusion model in terms of its target distribution $\rho$. To understand how $s_t$ changes in response to small perturbations of $\rho$, intuitively one would like to differentiate $s_t$ with respect to the probability measure $\rho$. However, it is not obvious how to compute this derivative in practice. In this section, we present a tractable formula for such a derivative with respect to \emph{additive} perturbations of $\rho$. This class of perturbations includes many cases of interest, such as the addition of new samples and the removal of existing samples from the training set.

Suppose that $\rho^\eta := (1- \eta) \rho + \eta \nu $ is a \emph{mixture} of two probability measures $\rho$ and $\nu$ supported on $\R^d$, and let $s^\eta_t : \R^d \rightarrow \R^d$ be the score function of a diffusion model with target distribution $\rho^\eta$ at time $t$. Differentiating $s^\eta_t$ with respect to $\eta$ and evaluating this derivative at $\bar{\eta}$ yields a \emph{function} $g^{\bar{\eta}}_t : \R^d \rightarrow \R^d$ describing how $s^\eta_t$ changes as one infinitesimally up-weights $\nu$ given initial weight $\bar{\eta}$.

The case $\bar{\eta}=0$ is of particular interest. For example, if $\bar{\eta} = 0$, then $g^{\bar{\eta}}_t$ describes how a score function trained on $\rho$ would vary as one introduces samples from $\nu$. On the other hand, to approximate how the score function of a diffusion model trained on $\rho$ would change in response to \emph{removing} training data lying in some region $\Omega \subseteq \R^d$, one would define $\nu := \rho_\Omega $, where $\rho_{\Omega}$ is the restriction of $\rho$ to $\Omega$, and consider $-g^{\bar{\eta}}_t$ evaluated at $\bar{\eta} = 0$.

\textbf{Our key result is the following theorem}, which provides a tractable closed-form expression for $g^{0}_t$:

\begin{theorem}[Sensitivity analysis for score functions]\label{thm:score-sensitivity}
    For $\eta \in [0,1]$, let $\rho^\eta := (1- \eta) \rho + \eta \nu $ be a mixture of probability measures $\rho$ and $\nu$ with compact support on $\R^d$. Let $\rho^\eta_t : \R^d \rightarrow \R$ and $s^\eta_t : \R^d \rightarrow \R^d$ be the density and score function, resp., of a diffusion model with target distribution $\rho^\eta$ at time $t \in [t_0, t_1)$. Then 
    the Fr\'{e}chet derivative in $L^2(\R^d, \rho^\eta_t)$ of the map $T_t(\eta) : \eta \mapsto s^\eta_t$ evaluated at $\bar{\eta}=0$
    is the function $g^0_t : \R^d \rightarrow \R^d$ defined by the formula:
    \begin{equation}\label{eq:score-sensitivity-analysis}
        g^0_t(z) = \frac{\nu_t(z)}{\rho_t(z)}\left(s^\nu_t(z) - s^\rho_t(z) \right),
    \end{equation}
    where $\nu_t(z), \rho_t(z)$ are the respective densities and $s^\nu_t(z),s^\rho_t(z)$ the respective scores at time $t$ of diffusion models with target measures $\nu, \rho$.
\end{theorem}

\begin{wrapfigure}[11]{r}{0.4\textwidth} \vspace{-.1in}
    \centering
    \includegraphics[width=\linewidth]{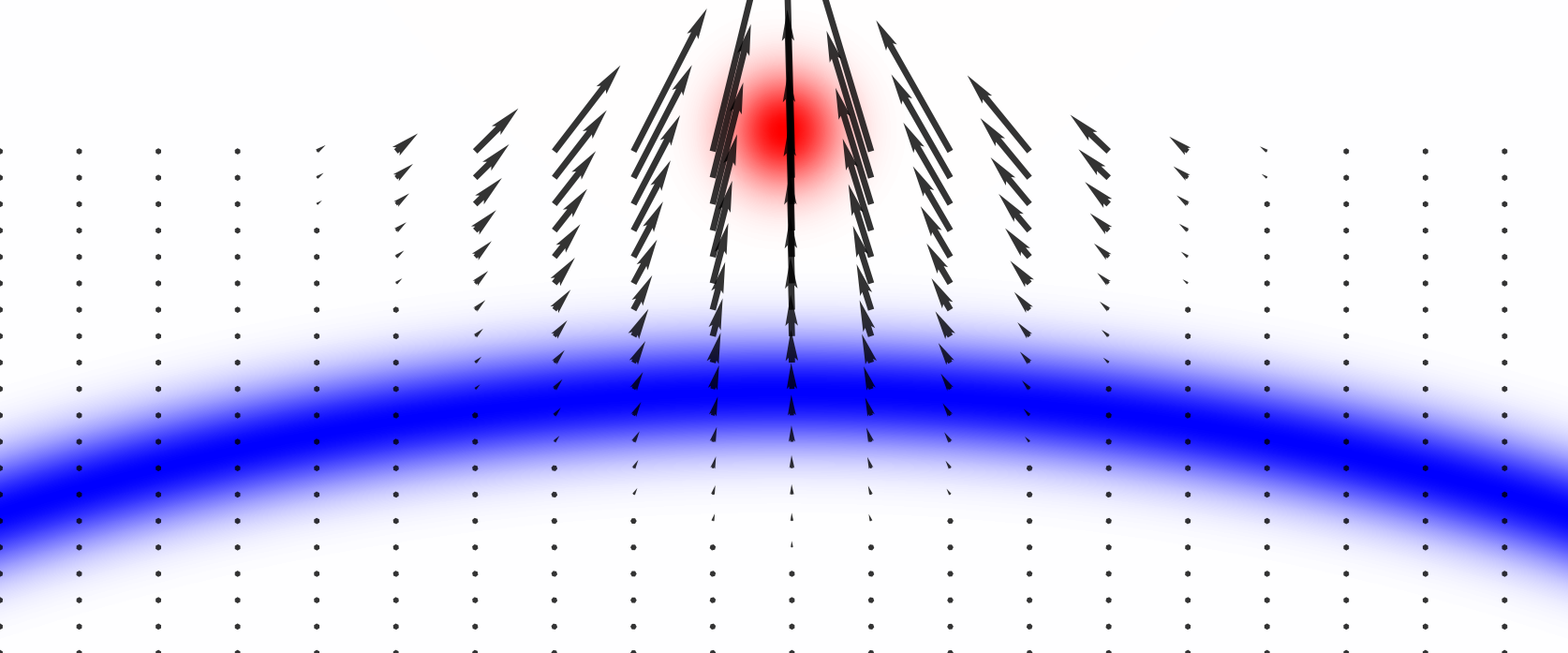}
    \caption{The score sensitivity $g^0_t$ is a vector field pointing away from the support of {\color{blue}$\rho$} and towards the support of {\color{red}$\nu$}.}
    \vspace{-10pt}
    \label{fig:score-sensitivity}
\end{wrapfigure}

We prove this result in Appendix \ref{pf:score-sensitivity}. Whereas \Eqref{eq:score-function} shows that there exists a well-defined map from a measure $\rho$ to the optimal score function $s_t$ of a diffusion model with $\rho$ as its target, \Eqref{eq:score-sensitivity-analysis} now provides a formula for the \emph{directional derivative} of this map in the direction of $\nu - \rho$. In Figure \ref{fig:score-sensitivity}, we depict an instance of this directional derivative $g^0_t$ when $\rho$ is supported on a curve in 2D and $\nu$ is a Gaussian measure centered just off the curve. $g^0_t$ is a vector field pointing away from the support of $\rho$ and towards the support of $\nu$; on account of the $\nicefrac{\nu_t(z)}{\rho_t(z)}$ scaling factor, $\|g^0_t(z)\|_2$ is large at points $z$ that are closer to the support of $\nu$ than to the support of $\rho$.

In a typical use case, $\rho_t$ is the distribution at time $t$ of a diffusion model trained on $\rho$, and $\nu = \frac{1}{K}\sum_{k=1}^K \delta_{x_k}$ is the empirical distribution on $K$ samples $x_k$ that one wishes to add or remove from $\rho$; if $K=1$, we recover the \emph{influence function} of $T_t$ \citep{hampel74}. In this setting, we may use \Eqref{eq:score-sensitivity-analysis} to compute $g_t(z)$ given \emph{only black-box access} to the score function $s^\rho_t(z)$ and the $K$ samples $x_k$. The density $\rho_t(z)$ of the diffusion model can be computed from its score using the continuous change of variables (CCoV) formula \citep{song2021scorebased}, and since $\nu_t$ is a mixture of Gaussians when $\nu$ is an empirical distribution, its density and score function can be computed in closed form or efficiently approximated using techniques from \citet{scarvelis2025closedform}. For perturbation sets $S = \{x_k\}_{k=1}^K$ of moderate size, the cost of evaluating \Eqref{eq:score-sensitivity-analysis} is dominated by the cost of the density computations.

Theorem \ref{thm:score-sensitivity} shows how to tractably estimate the response of a pre-trained score function to additive perturbations in its target distribution. However, in practice, we are typically interested in how the \emph{samples} generated by a diffusion model would change in response to perturbing its target distribution. Because samples are obtained by solving an ODE or SDE determined by the score function, \Eqref{eq:score-sensitivity-analysis} should  provide enough information to estimate the sensitivity of model samples to additive perturbations of $\rho$. We show this to be the case in the following section, using the adjoint method to obtain an analogous perturbation formula for a diffusion model's samples.

\subsection{Sensitivity analysis for model samples}\label{sec:sample-sensitivity}

Diffusion models generate samples from their target distribution $\rho$ by solving a stochastic differential equation (SDE) or an ordinary differential equation (ODE) whose \emph{drift} or \emph{velocity field}, respectively, depend on the score $s^\eta_t$. Because this dependence is typically simple, often consisting of an affine transformation of $s^\eta_t$, it is easy to differentiate the drift or velocity field with respect to $\eta$ given \Eqref{eq:score-sensitivity-analysis}. In this section, we will exploit this fact to compute the sensitivity of a diffusion model's samples with respect to additive perturbations.

\paragraph{ODE sampling.}

We begin with the simpler case of ODE sampling. \citet{song2021scorebased} show that one may sample a diffusion model by solving a \emph{probability flow ODE} (PF-ODE), whose initial condition is drawn from the Gaussian base distribution: $\frac{\textrm{d}z_t}{\textrm{d}t} = v^\eta_t(z_t)$ with $z_0 \sim \mathcal{N}(0,I)$. Because the Lipschitz constant of $s^\eta_t$ -- and consequently $v^\eta_t$ -- may blow up as $t \rightarrow t_1$, we follow a common convention from the theory of diffusion models and truncate integration of $v^\eta_t$ at some $\tilde{t}_1 < t_1$ \citep{debortoli2022convergence}. This convention aligns with typical diffusion model sampling schemes, which return samples at some time $\tilde{t}_1$ slightly earlier than the theoretical sampling interval endpoint $t_1$.

If one further assumes that the target distributions $\mu, \nu$ are compactly supported on $\R^d$, then a typical $v^\eta_t(z)$ will be globally Lipschitz for $z \in \R^d$ and $t \in [t_0, \tilde{t}_1]$. \citet[Theorem 3.2]{khalil2002nonlinear} then shows that there exists a unique solution to the PF-ODE for any initial condition $z_0 \in \R^d$. This allows us to define a \emph{solution map} $\Phi_s^\eta(z_0) : \R^d \rightarrow \R^d$ that maps an initial condition $z_0 \in \R^d$ to the unique solution at time $s \in [t_0, \tilde{t}_1]$ of the initial value problem (IVP) defined by $v^\eta_t$. Intuitively, $\Phi_s^\eta(z_0)$ maps an initial noise sample $z_0 \sim \mathcal{N}(0,I)$ to the sample's position at time $s$ along the diffusion model's sample path; at time $s = \tilde{t}_1$, this is simply a model sample.

We are interested in the derivative $\frac{\textrm d}{\textrm d \eta} \Phi_{\tilde{t}_1}^\eta(z_0)$ for fixed initial conditions $z_0$, which describes how the model sample $\Phi_{\tilde{t}_1}^\eta(z_0)$ generated from the Gaussian sample $z_0$ varies as one perturbs the target distribution $\mu$ in the direction of $\nu$. \citet[Section 3.3]{khalil2002nonlinear} shows that under certain regularity conditions, this derivative solves an ODE known as the \emph{sensitivity equation}. Defining $\psi_s := \frac{\textrm d}{\textrm d \eta} \Phi_{s}^\eta(z_0)$ and letting $z_s := \Phi_{s}^\eta(z_0)$ for $s \in [t_0,\tilde{t}_1]$ be a solution path for the PF-ODE, this equation is:
\begin{equation}\label{eq:ode-sensitivity}
    \frac{\textrm d}{\textrm d s} \psi_s = \frac{\textrm d}{\textrm d \eta} v^\eta_s(z_s) + J_z[v^\eta_s](z_s) \psi_s,
\end{equation}
\begin{wrapfigure}[11]{r}{0.4\textwidth} \vspace{-.25in}
    \centering
    \includegraphics[width=\linewidth]{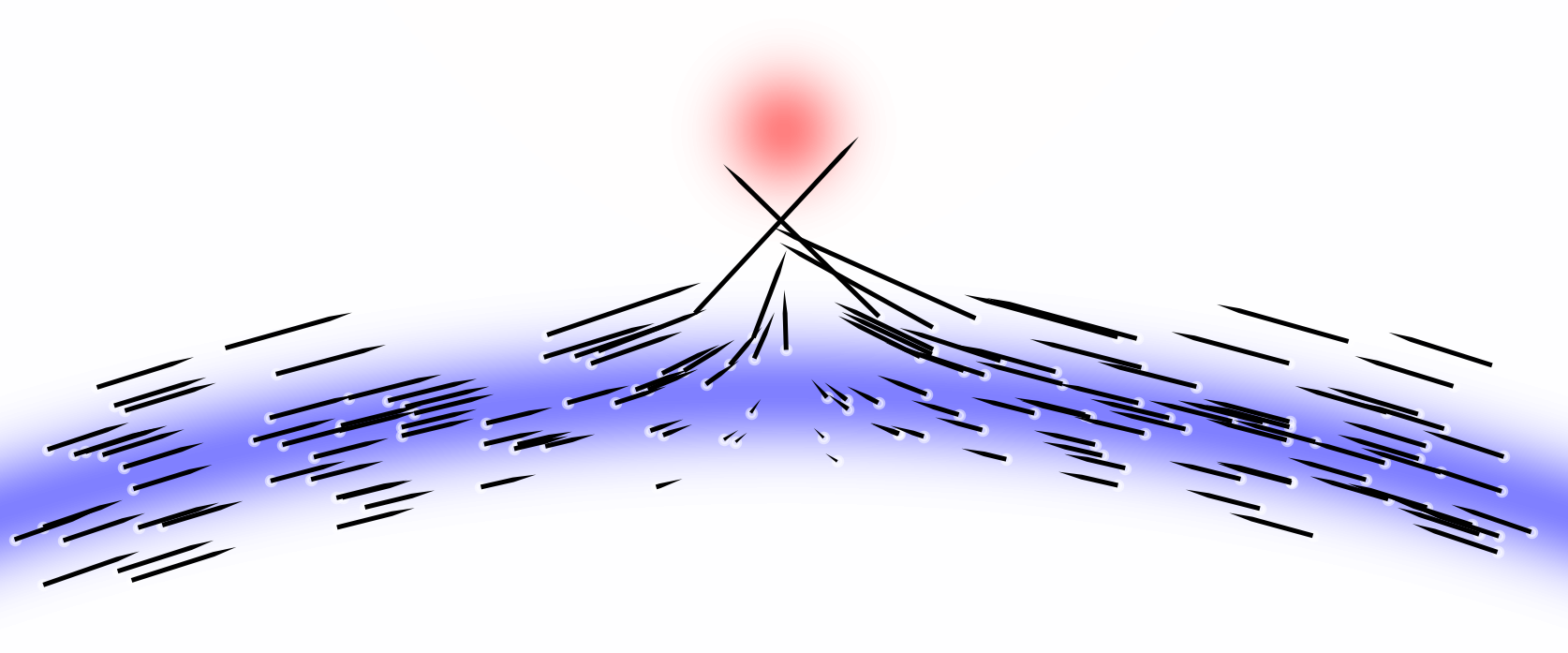}
    \caption{A solution to \Eqref{eq:ode-sensitivity} approximates the change in a diffusion model's samples as the target distribution {\color{blue}$\rho$} is perturbed in the direction of {\color{red}$\nu$}.}
    \label{fig:sample-sensitivity}
    \vspace{-10pt}
\end{wrapfigure}
where the initial condition is $\psi_{t_0} = 0$ and $J_z[v^\eta_s](z_s)$ denotes the spatial Jacobian of $v^\eta_s$ evaluated at $z_s$. A solution $\psi_{\tilde{t}_1} = \frac{\textrm d}{\textrm d \eta} \Phi_{\tilde{t}_1}^\eta(z_0)$ to \Eqref{eq:ode-sensitivity} approximates the change in a sample $z_{\tilde{t}_1} = \Phi_{\tilde{t}_1}^\eta(z_0)$ in response to additive perturbations of the target distribution $\rho$. Crucially, one may solve this IVP given black-box access to the score function $s^\eta_s$ and its spatial derivatives. To estimate how a sample $\Phi_s^\eta(z_0)$ generated from initial noise $z_0$ would change in response to perturbing $\rho$, one should (1) compute a sample path $z_t$ and model densities $\rho_t(z_t)$ by jointly integrating the PF-ODE and the CCoV formula, (2) evaluate \Eqref{eq:score-sensitivity-analysis} along the sample path, which also entails computing the density and score of the perturbation measure $\nu$, and (3) integrate \Eqref{eq:ode-sensitivity}, using autograd to compute the spatial Jacobian-vector products $J_z[v^\eta_s](z_s) \psi_s$. Figure \ref{fig:sample-sensitivity} depicts a solution to \Eqref{eq:ode-sensitivity} for when $\rho$ (in blue) is supported on a curve in 2D and $\nu$ (in red) is a Gaussian measure centered just off the curve.

\paragraph{SDE sampling.}

In practice, it is more common to sample a diffusion model by solving an SDE $\textrm d z_t = f^\eta_t(z_t) \textrm d t + g_t \textrm d W_t$, where $W_t$ denotes a Wiener process on $\R^d$. Only the drift coefficient $f^\eta_t$ depends on the score function $s^\eta_t$ and consequently on $\eta$; conversely, the \emph{diffusion coefficient} $g_t$ is independent of $\eta$. \citet[Theorem 3.3.2]{kunita2019stochastic} provides an analogous sensitivity analysis for the solution of an SDE whose coefficients depend on a parameter. Suppose the SDE has a unique solution and let $\Gamma^\eta_{s,\omega} : \R^d \rightarrow \R^d$ be the solution map sending an initial condition $z_0$ to the solution to the SDE at time $s \in [t_0, t_1]$ for some fixed realization $\omega$ of the Wiener process. \citet[Theorem 3.3.2]{kunita2019stochastic} shows that under certain regularity conditions, which are satisfied for typical drifts if one truncates the integration at $\tilde{t}_1 < t_1$, $\frac{\textrm d}{\textrm d \eta} \Gamma^\eta_{s,\omega}(z_0)$ satisfies a particular SDE for almost all $\omega$. Moreover, when the diffusion coefficient $g_t$ is independent of $\eta$ and the spatial variable, this differential equation is, in fact, deterministic and coincides with the sensitivity analysis for ODE sampling specified by \Eqref{eq:ode-sensitivity}. We may therefore use \Eqref{eq:ode-sensitivity} to approximate the change in a diffusion model's SDE samples in response to perturbations of its target distribution. In practice, one follows the recipe from the previous section on ODE sampling, but replaces the ODE sample path $z_t$ with an SDE sample path.

\section{Experiments}
\label{sec:experiments}

In this section, we empirically validate our sensitivity analysis for diffusion models. We initially study the impact of approximation error on our method's accuracy using synthetic examples where exact score functions and log-probabilities are available. We then experiment with neural diffusion models trained on image datasets and show that our sample sensitivities correlate with changes in model samples after retraining and after fine-tuning.


\subsection{First-order approximation for perturbed model samples}\label{sec:stepsize-hutch-val}

A solution $\psi_{\tilde{t}_1} = \frac{\textrm d}{\textrm d \eta} \Phi_{\tilde{t}_1}^\eta(z_0)|_{\eta = 0}$ to \Eqref{eq:ode-sensitivity} estimates the change in a diffusion model's samples in response to an additive perturbation of the target distribution. Using this sensitivity analysis, one may compute a first-order approximation of the samples a diffusion model trained on $\rho^0$ would have generated if its target distribution were $\rho^{\bar{\eta}}$: 
\begin{equation}\label{eq:first-order-sample-approx}
    \Phi_{\tilde{t}_1}^{\bar{\eta}}(z_0) \approx \Phi_{\tilde{t}_1}^0(z_0) + \bar{\eta} \frac{\textrm d}{\textrm d \eta} \Phi_{\tilde{t}_1}^\eta(z_0)\Big|_{\eta = 0}.
\end{equation}
Taylor's theorem states that this approximation converges at rate $o(\bar{\eta})$. However, because we estimate the derivative by solving a differential equation \plaineqref{eq:ode-sensitivity}, numerical error may render this approximation inaccurate for practical step sizes. Furthermore, evaluating \Eqref{eq:score-sensitivity-analysis} requires the density $\rho_t$ of a score model, typically approximated using the CCoV formula. In practice, this formula involves a noisy estimate of $\textrm{div}(v_t)$ using Hutchinson's trace estimator, introducing numerical error. In this section, we use synthetic data for which exact score functions and log-densities are available to study the impact of numerical error in ODE integration and density estimation on the convergence of our linear approximation to a perturbed model's samples. Appendix \ref{sec:first-order-exper-details} provides implementation details.

\paragraph{Effect of step size.}

In this experiment, we analyze the impact of step size on the convergence of our first-order approximation to the perturbed model's samples when solving \Eqref{eq:ode-sensitivity} using a forward Euler scheme. We let the initial target measure $\rho$ be an equally-weighted mixture of well-separated Gaussians on $\R^{100}$. This multimodal, high-dimensional target distribution simulates some of the pathologies of real-world data, and for all $t$, the distribution of the resulting diffusion model $\rho_t$ is a mixture of Gaussians, whose score and density are available in closed form. We perturb $\rho$ in the direction of a second Gaussian distribution $\nu$, which ensures that the perturbed target $\rho^{\bar{\eta}} = (1-\bar{\eta})\rho +\bar{\eta} \nu$ remains a mixture of Gaussians.

We obtain sample paths $z_t$ for $\rho_t$ and $\rho^{\bar{\eta}}_t$ by numerically integrating the PF-ODE and variance-preserving SDE using a forward Euler scheme and an Euler-Maruyama scheme, resp., with several step sizes, and exactly compute the Gaussian mixture densities $\rho_t(z_t)$ along each sample path. We then integrate \Eqref{eq:ode-sensitivity} using the same forward Euler scheme to obtain the sensitivities $\frac{\textrm d}{\textrm d \eta} \Phi_{\tilde{t}_1}^\eta(z_0)|_{\eta = 0}$ of samples from the initial target to perturbations in direction $\nu$. \Eqref{eq:first-order-sample-approx} suggests that we may use these sensitivities to estimate the final positions of samples from the perturbed model $\rho^{\bar{\eta}}$, and Taylor's theorem implies that the error $R(\bar{\eta})$ of this estimate is $o(\bar{\eta})$, so $\nicefrac{R(\bar{\eta})}{\bar{\eta}} \rightarrow 0$ as $\bar{\eta} \rightarrow 0$. We compute $R(\bar{\eta})$ for a variety of $\bar{\eta} \in [0,1]$ and verify that this rate holds in practice.

\begin{figure*}[t]
\centering
\subfigure{\includegraphics[scale=0.18]{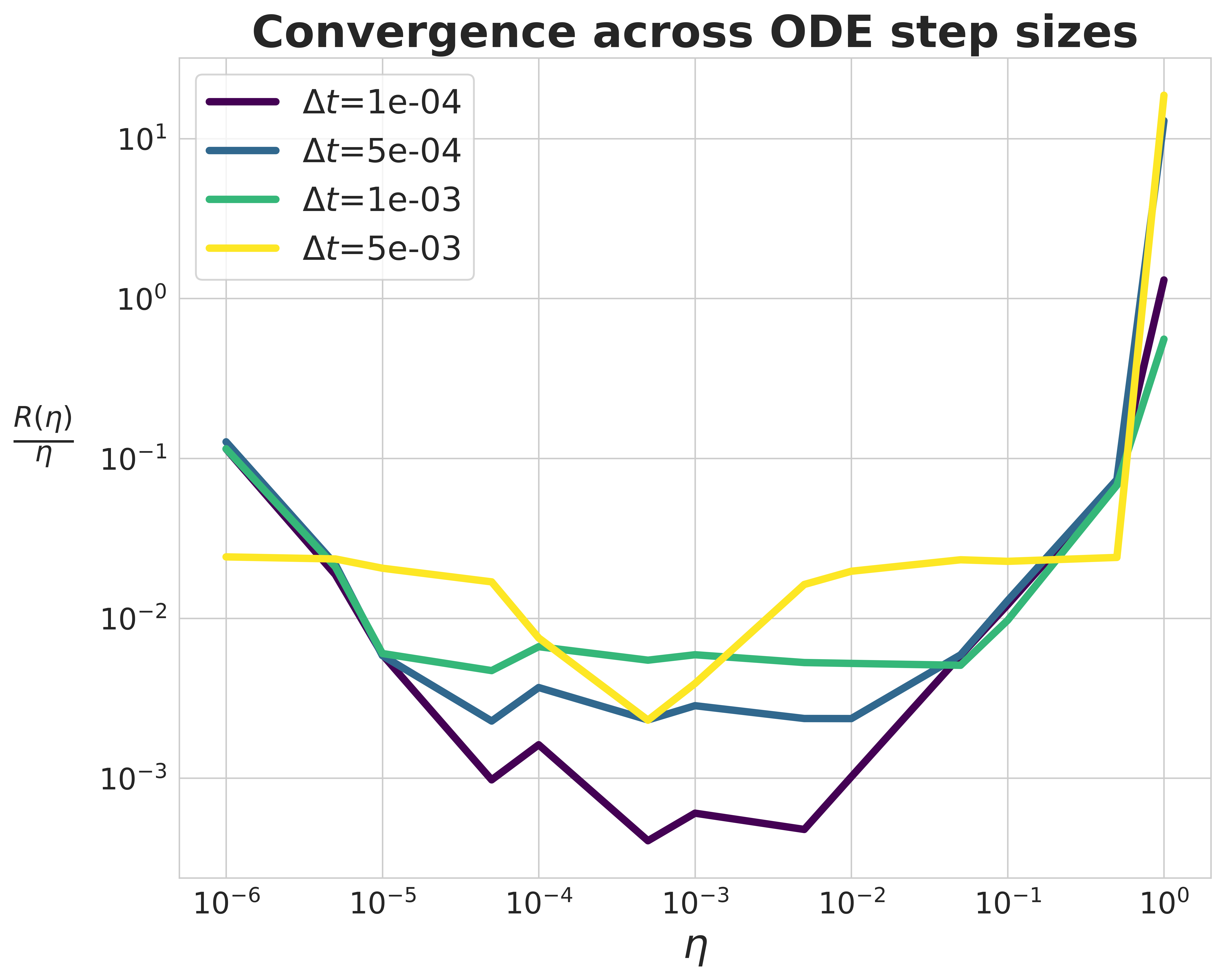}}
\subfigure{\includegraphics[scale=0.18]{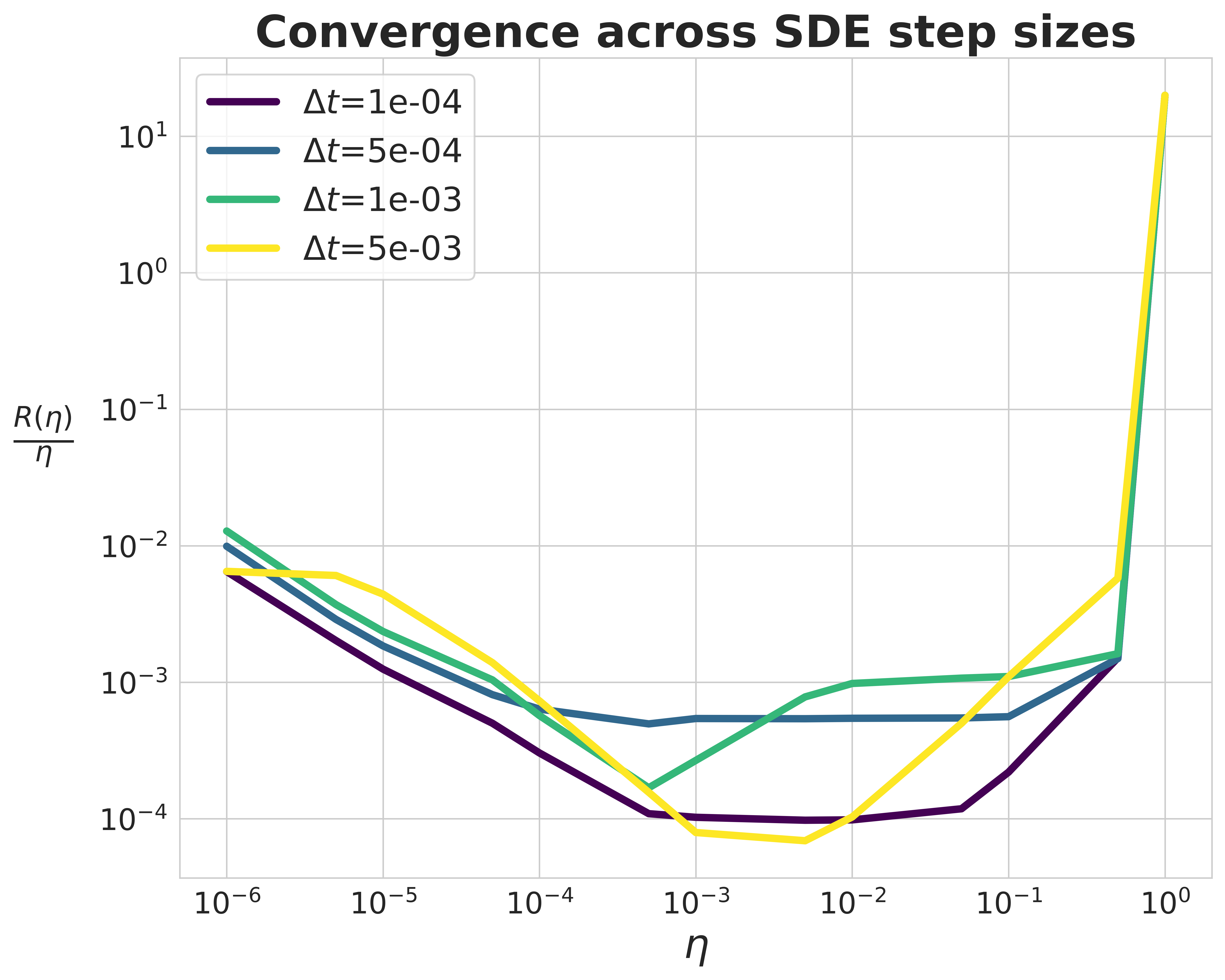}}
\subfigure{\includegraphics[scale=0.23]{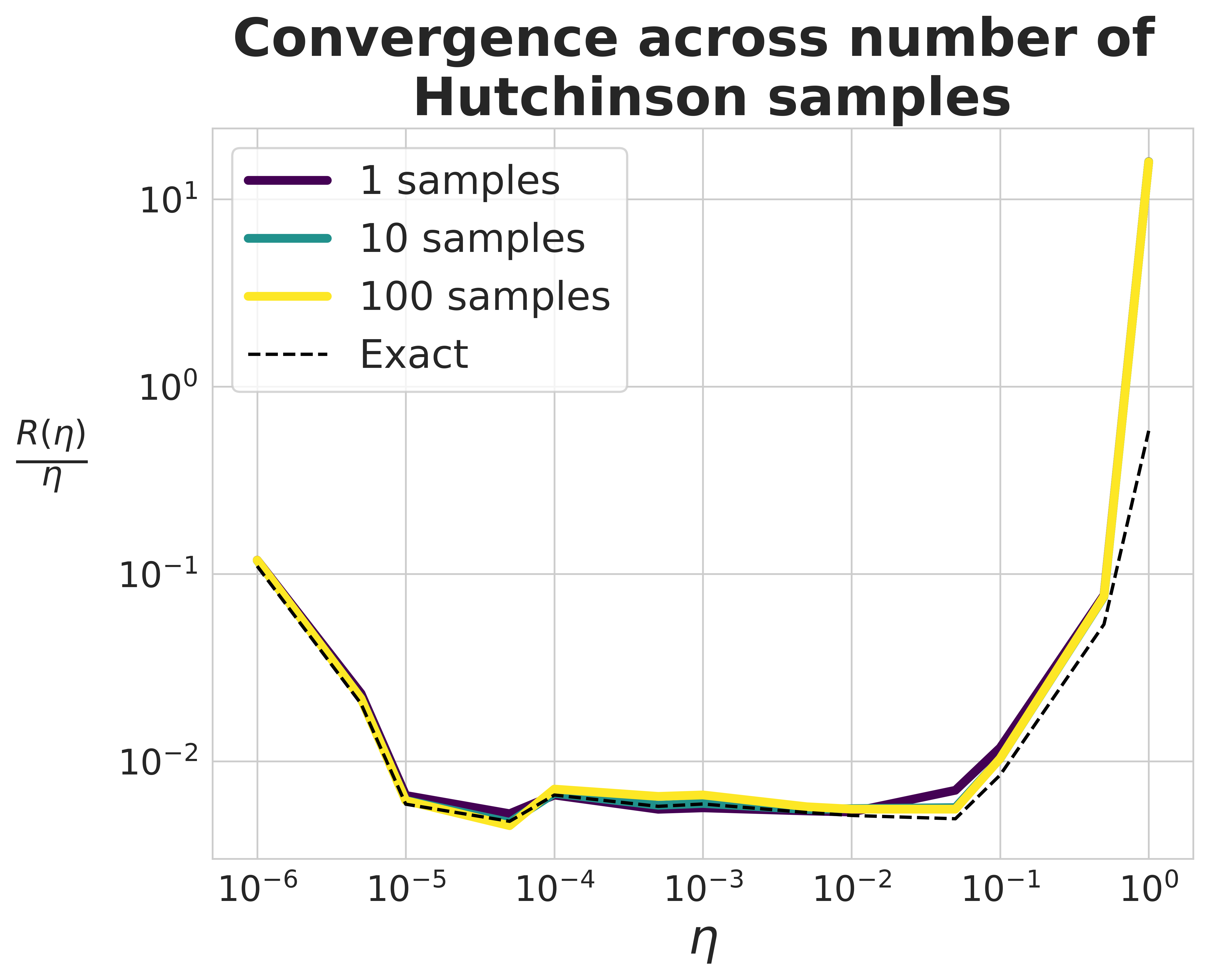}}
\vspace{-.15in}
\caption{The approximation error of our first-order approximation to a perturbed model's samples decays at rate $o(\bar{\eta})$ for a variety of ODE step sizes (left) and SDE step sizes (center), and this rate is robust to noise from Hutchinson's estimator (right).}\label{fig:ode-sde-error}
\vspace{-10pt}
\end{figure*}

The left and center panels of Figure \ref{fig:ode-sde-error} depicts the results of this experiment for ODE and SDE sample paths. Linearly approximating samples from a perturbed target measure $\rho^{\bar{\eta}}$ using our sample sensitivity analysis \plaineqref{eq:ode-sensitivity} is accurate within $o(\bar{\eta})$ for a variety of step sizes and $\bar{\eta}$. For very small values of $\bar{\eta}$, $\nicefrac{R(\bar{\eta})}{\bar{\eta}}$ plateaus and begins to increase again. This may reflect a noise floor in the accuracy of model samples $\Phi_{\tilde{t}_1}^{\bar{\eta}}(z_0)$ and $\Phi_{\tilde{t}_1}^0(z_0)$, which are themselves computed by numerically integrating an ODE or SDE.


\paragraph{Effect of Hutchinson's estimator.}

In the previous experiment, we exactly computed the model densities $\rho_t(z)$ that appear in the score sensitivity analysis \plaineqref{eq:score-sensitivity-analysis}. This was possible because we selected the target distribution $\rho$ to ensure that the densities $\rho_t(z)$ are available in closed form. In practice, one trains a neural network to approximate a diffusion model's score function $\nabla \log \rho_t(z)$ and obtains model densities via the continuous change of variables (CCoV) formula $\frac{\textrm{d} \log \rho_t(z_t)}{\textrm{d}t} = - \textrm{tr}(J_{z_t}[v^\eta_t](z_t))$. To avoid materializing a large Jacobian, one typically employs Hutchinson's trace estimator $\textrm{tr}(A) = \E[\epsilon^\top A \epsilon]$ to approximate the RHS, whose estimates are noisy due to the small number of $\epsilon$ samples used to compute the estimator. To study the impact of this noise on the convergence of \Eqref{eq:first-order-sample-approx}, we repeat the previous experiment with step size $10^{-3}$ while estimating the model densities $\rho_t(z)$ using the CCoV formula with a varying number of Hutchinson $\epsilon$ samples.

We depict the results of this experiment in the right panel of Figure \ref{fig:ode-sde-error}, where the scaled remainders $\nicefrac{R(\bar{\eta})}{\bar{\eta}}$ arising from using exact densities $\rho_t$ are represented by the dashed line, and we experiment with 1, 10, and 100 $\epsilon$ samples in Hutchinson's trace estimator. Our method's convergence rate is robust to noise in Hutchinson's trace estimator, with even a single draw of $\epsilon$ achieving nearly the same approximation error as the exact densities for all but the largest values of $\eta$.

\subsection{Stability of sample sensitivity under score approximation error}\label{sec:stability-neural-scores}

\begin{wrapfigure}{h}{0.5\textwidth} 
\vspace{-10pt}
    \centering
    \includegraphics[width=\linewidth]{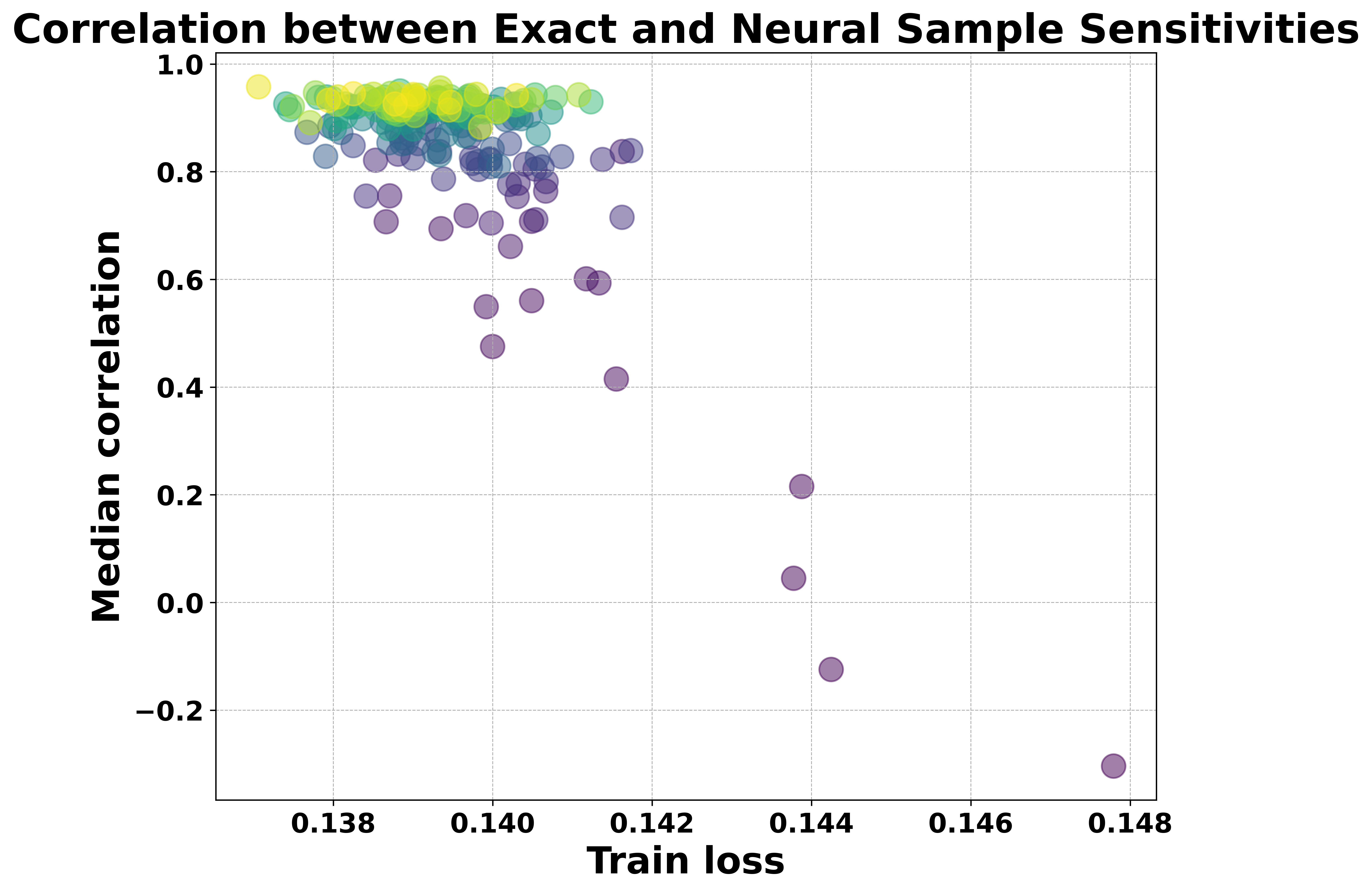}
    \caption{The correlation between sample sensitivities for an exact diffusion model and its neural approximation rises rapidly as the training loss falls. Points are colored from purple to yellow according to the training step.}
    \label{fig:sample-sensitivity-correlations}
    \vspace{-15pt}
\end{wrapfigure}

Section \ref{sec:stepsize-hutch-val} showed that one may approximate perturbed samples $\Phi_{\tilde{t}_1}^{\bar{\eta}}(z_0)$ using our sensitivity analysis formula \plaineqref{eq:ode-sensitivity} and recover the expected $o(\bar{\eta})$ convergence rate in spite of errors from numerical integration and Hutchinson's estimator. To isolate the effects of these errors, we used the exact score $\nabla \log \rho_t$ of the mixture of Gaussians $\rho_t$ throughout our computations. In practice, however, one typically \emph{learns} this score function by training a neural network to optimize a score-matching objective. Here, we show that our sample sensitivity analysis \plaineqref{eq:ode-sensitivity} is stable to error from neural approximations of the true score function.

In this experiment, our initial target measure $\rho$ is a mixture of well-separated Gaussians on $\R^{10}$, and we perturb $\rho$ in the direction of a Gaussian measure $\nu$. Instead of evaluating the score of $\rho_t$ in closed form as in Section \ref{sec:stepsize-hutch-val}, we train a neural network to approximate this score function. We fix the base samples $z_0 \sim \rho_0$ and evaluate the sensitivity of samples from the exact diffusion model $\rho_t$ and its neural approximation every 1000 training steps. We discretize all ODEs using forward Euler and use Hutchinson's estimator to estimate the model densities $\rho_t(z)$. We measure the median correlation between the exact and approximate sample sensitivities and compare it to the value of the score-matching loss at that training step. Appendix \ref{sec:stability-exper-details} provides further implementation details.

Figure \ref{fig:sample-sensitivity-correlations} shows the relationship between the training loss and the median correlation between the exact and approximate sample sensitivities at each model sample. The points are colored from purple to yellow according to the training step at which the loss and sensitivities were measured; for clarity, we omit the first two measurements where the training loss is very large. Our sample sensitivity analysis is stable to approximation error in the score function, with correlations between sample sensitivities computed with the exact diffusion model and its neural approximation rising rapidly as the training loss falls. This shows that our formulas may provide useful information even when an exact score function is replaced by a neural approximation. In the following section, we build on this observation by showing that our sample sensitivities are correlated with the direction of change in a diffusion model's samples after retraining on a perturbed training distribution.

\subsection{Image datasets}\label{sec:image-val}

\paragraph{Predicting change in model samples after retraining.}

In the previous section, we used synthetic data from a mixture of Gaussians to study the robustness of our sensitivity analysis to various sources of numerical error. In practice, diffusion models are trained on large datasets of images, with training often stopped well before convergence to prevent memorization \citep{favero2025}. In this section, we demonstrate that our sample sensitivities $\psi_{\tilde{t}_1}$ are correlated with differences between an image diffusion model's samples before and after retraining on a perturbed target distribution. We experiment with UNet-based diffusion models trained on a mixture of the MNIST and Typography-MNIST (TMNIST) datasets \citep{tmnist22} and on the CelebA dataset \citep{liu2015faceattributes}.

For each dataset, we train a base model and a perturbed model whose target distribution $\rho^{\bar{\eta}}$ is a mixture of the base model's target distribution and the empirical measure on a set of new samples $S$. We employ mixture weights $\bar{\eta}=0.1$ and $1-\bar{\eta}=0.9$, resp. For our MNIST experiment, the new samples $S$ are drawn from TMNIST, and for our CelebA experiment, $S$ consists of samples with a large CLIP score for ``a photo of an old man.'' We integrate the PF-ODE to obtain model samples from $\rho^0$ and $\rho^{\bar{\eta}}$, and also integrate \Eqref{eq:ode-sensitivity} with the perturbation measure $\nu$ set to the empirical distribution over $S$ to estimate the sensitivity of the base model's samples to upweighting $S$. We compare the sample sensitivities $\frac{\textrm d}{\textrm d \eta} \Phi_{\tilde{t}_1}^\eta(z_0)|_{\eta=0}$ to the difference $\Phi_{\tilde{t}_1}^{\bar{\eta}}(z_0) - \Phi_{\tilde{t}_1}^{0}(z_0)$ between PF-ODE samples from the perturbed and base model given the same initial noise. This measures how much our sample sensitivity analysis predicts actual changes in model samples after retraining on the perturbed target distribution $\rho^{\bar{\eta}}$. Appendix \ref{sec:image-exper-details} provides further implementation details.

\begin{figure*}[h]
\centering
\subfigure{\includegraphics[scale=0.2]{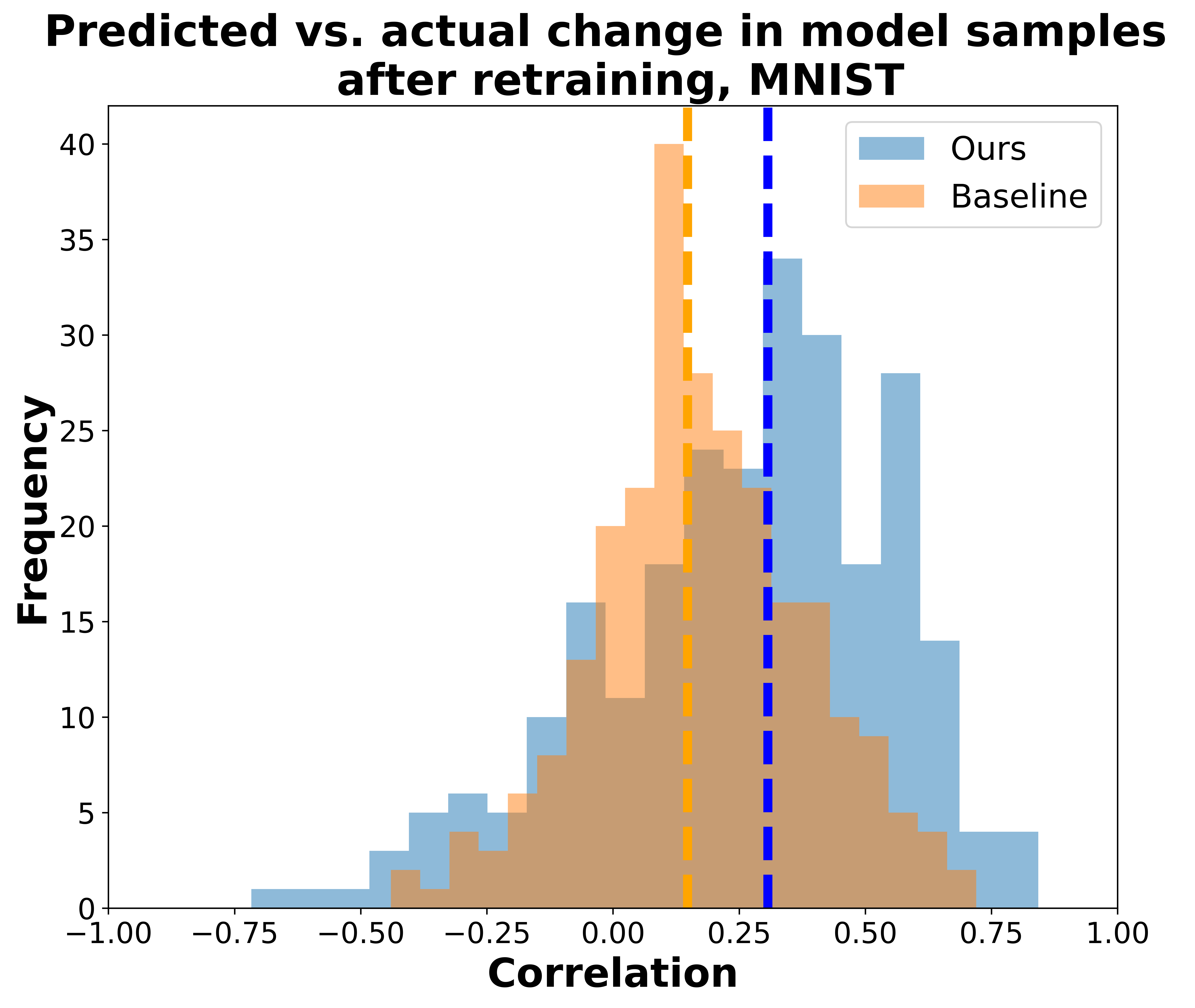}}\quad
\subfigure{\includegraphics[scale=0.2]{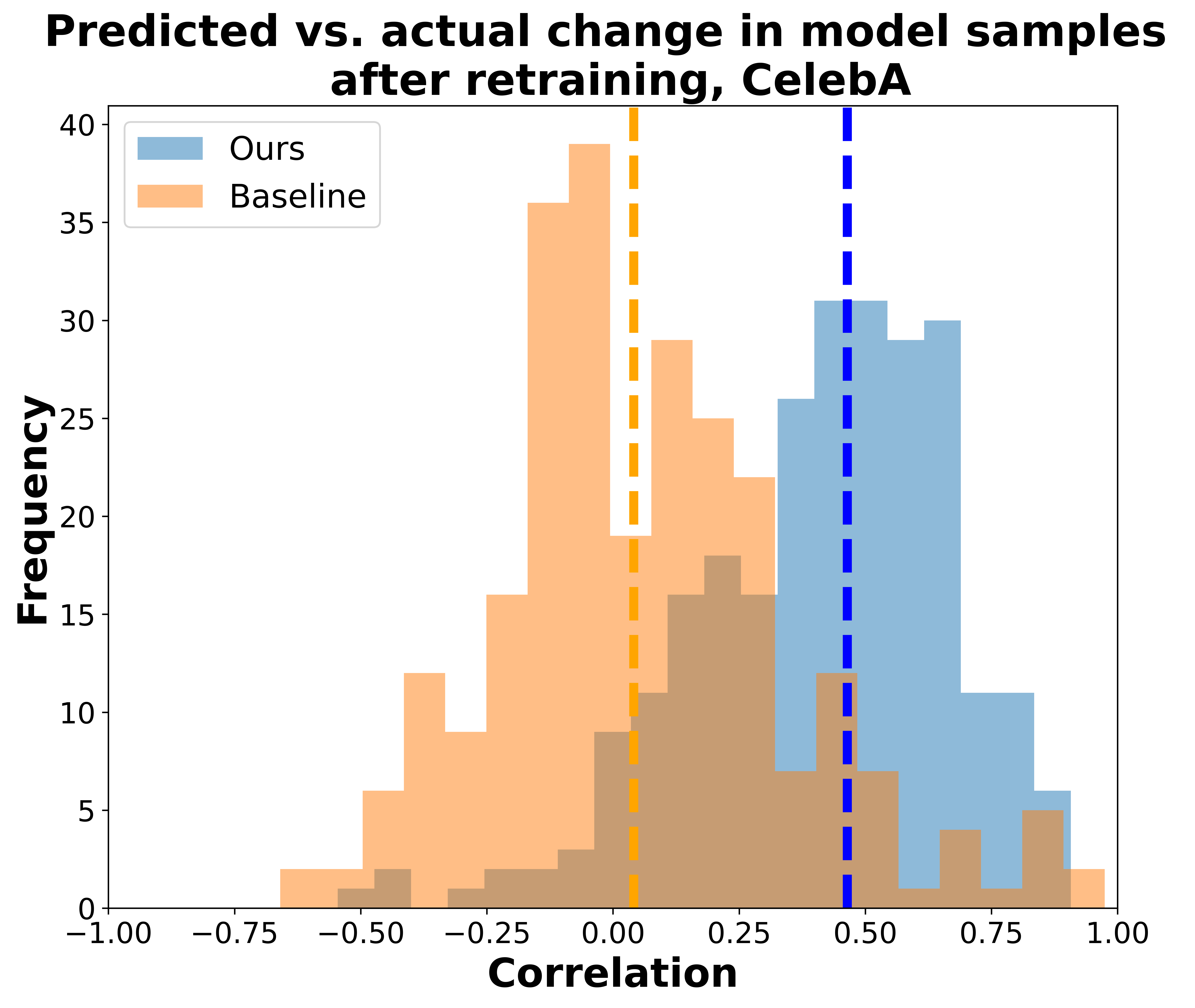}}
\vspace{-.15in}
\caption{Correlations between predicted and actual change in model samples after retraining on a perturbed dataset. Our sample sensitivity analysis (blue) outperforms an optimal transport baseline (orange), achieving a median correlation (dashed blue line) of 0.46 on CelebA and 0.31 on MNIST.}\label{fig:image-sample-sensitivity-retrain}
\end{figure*}

Figure \ref{fig:image-sample-sensitivity-retrain} depicts histograms of the correlations between our sample sensitivities and the actual change in model samples. As a baseline, we also compute the entropic optimal transport (OT) coupling \citep{cuturi2013} between the base model samples $\Phi_{\tilde{t}_1}^{0}(z_0)$ and the target distribution for the perturbed model and use the resulting transport rays as predicted directions of change in the model samples after retraining. Our sample sensitivity scores correlate with actual changes in model samples after retraining on $\rho^{\bar{\eta}}$ and substantially outperform the OT baseline, achieving a median correlation of 0.46 on CelebA and 0.31 on MNIST, compared 0.04 and 0.15, resp., for the OT baseline.

\paragraph{Predicting change in model samples after fine-tuning.}

The previous experiment shows that our sample sensitivities $\frac{\textrm d}{\textrm d \eta} \Phi_{\tilde{t}_1}^\eta(z_0)|_{\eta=0}$ correlate with changes in model samples after retraining on a perturbed target distribution. We will now show that our sample sensitivities are more strongly predictive of changes in model samples after \emph{fine-tuning} on new training samples $S$. We use the same base models and the same $S$ as in the previous experiment, but fine-tune on $S$ rather than retraining from scratch on the mixture distribution $\rho^{\bar{\eta}}$.

\begin{figure*}[h]
\centering
\subfigure{\includegraphics[scale=0.2]{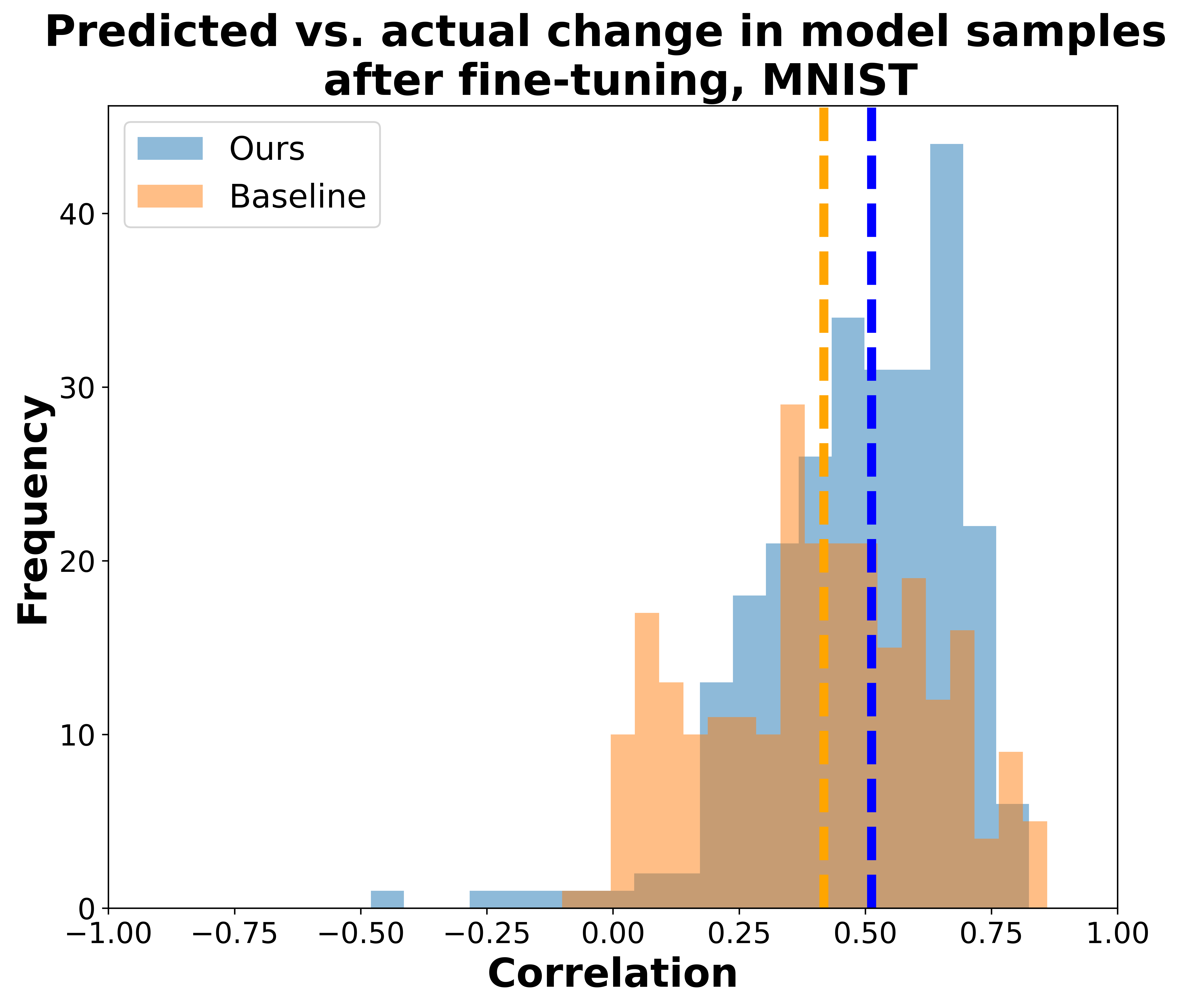}}\quad
\subfigure{\includegraphics[scale=0.2]{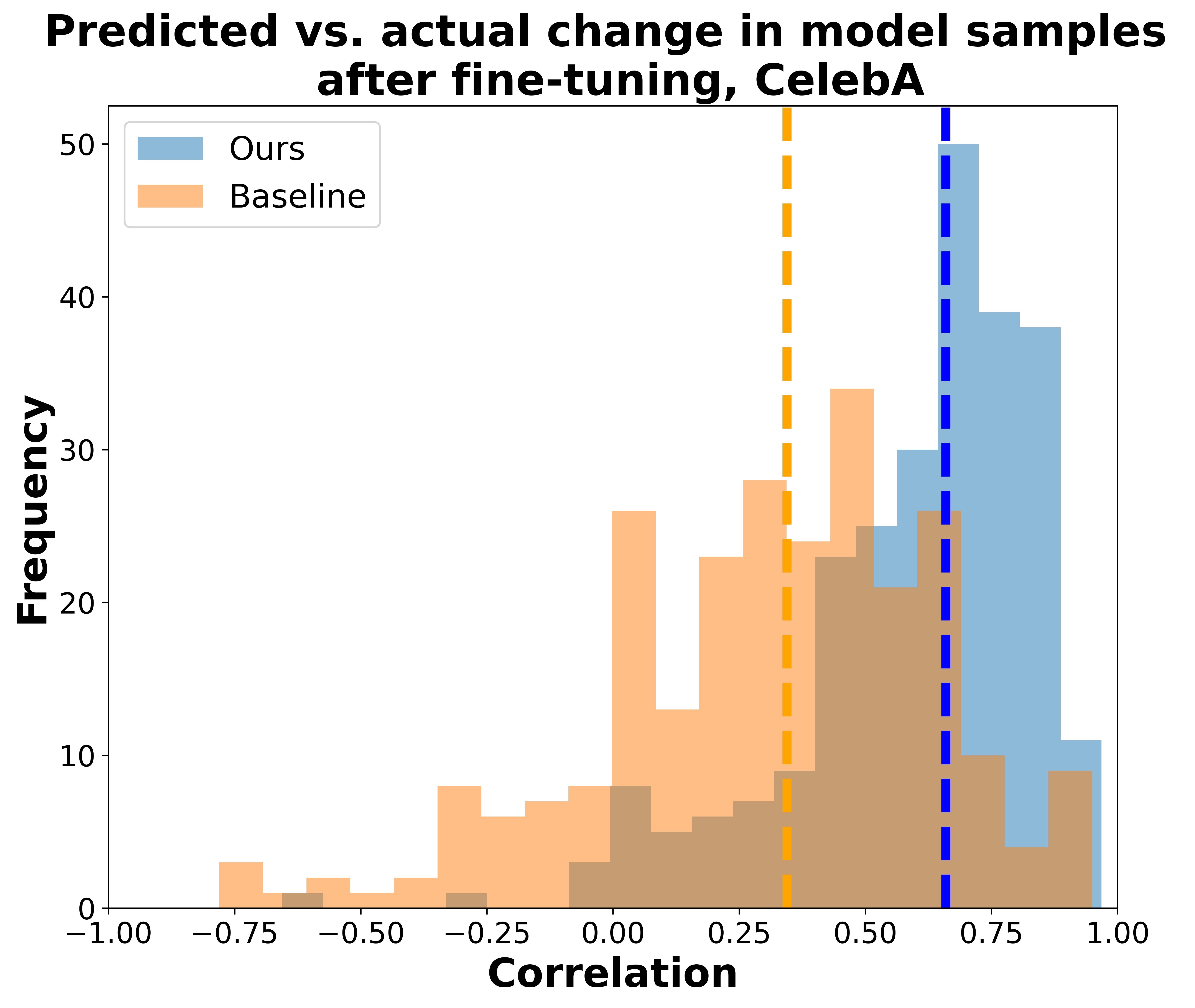}}
\vspace{-.15in}
\caption{Our sample sensitivities (blue) are correlated with changes in model samples after fine-tuning, and continue to outperform an optimal transport baseline (orange).}\label{fig:image-sample-sensitivity-finetune}
\end{figure*}

\begin{figure*}[h]
\centering
\subfigure{\includegraphics[scale=0.20]{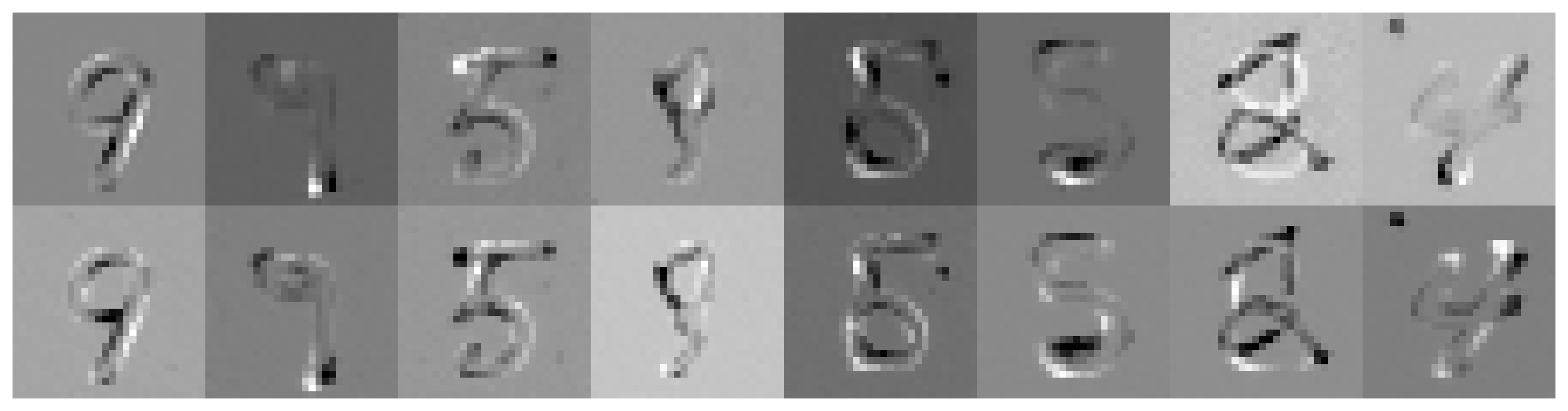}}\quad
\subfigure{\includegraphics[scale=0.20]{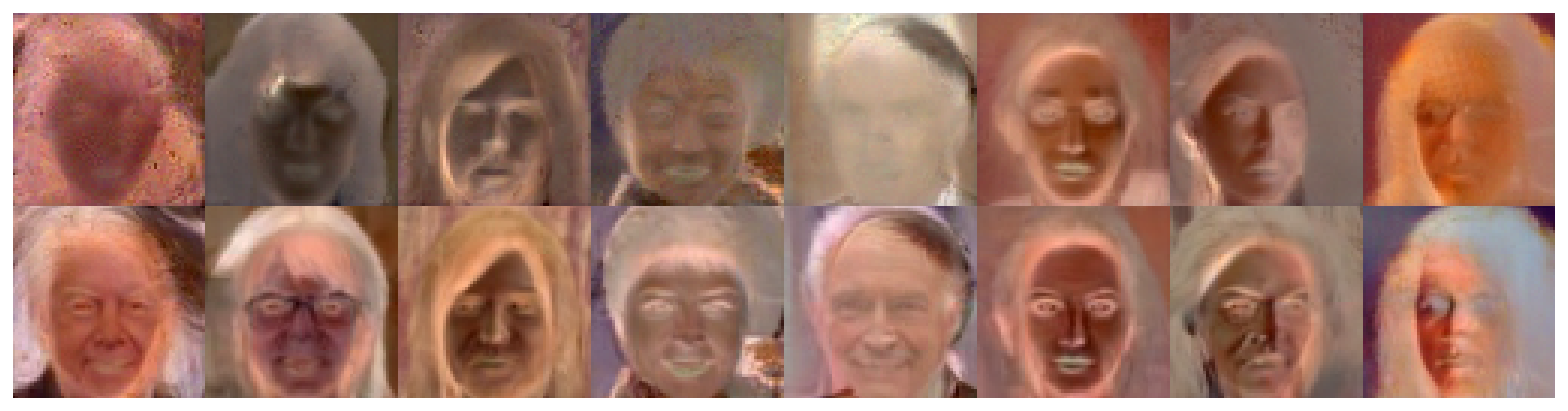}}
\vspace{-.15in}
\caption{Our sample sensitivities (top rows) predict changes in model samples after fine-tuning (bottom rows). The models that generated the samples on the left were trained on MNIST and fine-tuned on TMNIST. The models that generated the samples on the right were trained on CelebA and fine-tuned on a subset of photos with a large CLIP score for ``a photo of an old man.''}\label{fig:image-finetuning-perturbation-viz}
\vspace{-10pt}
\end{figure*}

We depict histograms of the correlations between our sample sensitivities and actual change in model samples after fine-tuning in Figure \ref{fig:image-sample-sensitivity-finetune}. We use the same entropic OT baseline as in the previous experiment, but compute transport rays between the base model samples and the samples $S$ on which we fine-tuned. Both our sample sensitivities and the OT baseline are better correlated with actual change in model samples after fine-tuning, but our method continues to outperform the baseline, achieving a median correlation of 0.66 on CelebA and 0.51 on MNIST, compared to 0.34 and 0.42, resp., for the baseline. We visually compare our sample sensitivities to actual changes in model samples after fine-tuning in Figure \ref{fig:image-finetuning-perturbation-viz}, which shows that our sample sensitivity analysis can provide coarse predictions of how a diffusion model's samples might change after fine-tuning. We provide further illustrations of our sample sensitivity analysis in Appendix \ref{sec:sample-sensitivity-viz}.

\section{Discussion}\label{sec:conclusion}

Understanding a diffusion model's dependence on its training data is a critical challenge in machine learning. In general, one would expect the relationship between a large model and its training data to be complex and difficult to estimate. This paper shows that it is not only possible to compute directional derivatives of the map from a training distribution $\rho$ to its optimal score function $s_t$, but that this computation is (a) surprisingly cheap, costing roughly as much as sampling a model and computing log-probabilities along the sample path, and (b) requires only black-box access to the score function. One can then leverage this simple formula to estimate how a diffusion model's samples change in response to perturbations to its target distribution before retraining or fine-tuning on new data. We propose several future directions for this line of work.

Throughout this paper, we perturbed a diffusion model's target measure with empirical measures over finite samples. This need not be the case: Our score sensitivity formula \plaineqref{eq:score-sensitivity-analysis} holds for any compactly-supported perturbation measure $\nu$, and it can be implemented in practice for any sequence of measures $\nu_t$ provided we can access their scores and densities. For instance, $\nu_t$ can be a second diffusion model, in which case \Eqref{eq:score-sensitivity-analysis} resembles the formula for classifier-free guidance (CFG) \citep{ho2022classifier} with time- and spatially-varying weights. Future work might interpret CFG in light of our sensitivity analysis and design new guidance schedules based on this formula.

By composing a model's ODE sampling solution map with a text-conditioned classifier and applying our sensitivity formulas, one might also use our method to estimate how the likelihood that a diffusion model's samples match a prompt changes as one perturbs the training set. This would allow users to attribute a model's qualitative behavior to subsets of training samples and use this information to curate the training set to steer a diffusion model's behavior in a particular direction.

Finally, \citet{kadkhodaie2024generalization} find empirically that a diffusion model's sampling map is often insensitive to changes in its training set, and \citet{favero2025} clarify that this behavior is controlled by the number of training iterations, with models becoming increasingly sensitive to dataset perturbations throughout training. Our sample sensitivity formula \plaineqref{eq:ode-sensitivity} \emph{quantifies} this dependence and may serve as a valuable tool for future work on generalization in diffusion models.

\subsubsection*{Acknowledgments}
The MIT Geometric Data Processing Group acknowledges the generous support of Army Research Office grant W911NF2110293, of National Science Foundation grants IIS2335492 and OAC2403239, from the CSAIL Future of Data and FinTechAI programs, from the MIT–IBM Watson AI Laboratory, from the Wistron Corporation, from the MIT Generative AI Impact Consortium, from the Toyota–
CSAIL Joint Research Center, and from Schmidt Sciences.

Christopher Scarvelis gratefully acknowledges the support of a Schwarzman College of Computing Future Research Cohort Fellowship, from the MIT-Google Program for Computing Innovation, and from an Amazon Research Award.



\bibliography{iclr2026_conference}
\bibliographystyle{iclr2026_conference}

\appendix

\section{Proofs}\label{sec:proofs}

\subsection{Proof of Theorem \ref{thm:score-sensitivity}}\label{pf:score-sensitivity}

We will prove this theorem in two parts. 
We will first show that $\frac{\partial}{\partial\eta} s^\eta_t(z) \Big |_{\eta = 0} = g_t(z)$ at any fixed $z \in \R^d$. This shows that $g_t(z)$ is the pointwise derivative of $s^\eta_t$ evaluated at $\eta=0$ at any $z \in \R^d$. We will then extend this pointwise argument to the space of functions by using the dominated convergence theorem (DCT) to prove that $g_t$ is the Fr\'{e}chet derivative in $L^2(\R^d, \rho^0_t)$ of the map $T(\eta) : \eta \mapsto s^\eta_t$.

\subsubsection{$g_t(z)$ is the pointwise derivative of $s^\eta_t$ at $\eta=0$}

In this part of the proof, we will rely heavily on \citet[Lemma 1]{mlodozeniec2025influence}. A version of their lemma adapted to our setting states the following:

\begin{lemma}\label{thm:mlo-lemma}
    Let $\Ell : \R \times \R^d \rightarrow \R$ be a $C^2$ function of additive form $\Ell(\eta, s_z) := \Ell_1(s_z) + \eta \Ell_2(s_z)$, and suppose that the map $s_z \mapsto \Ell(\eta,s_z)$ is strictly convex for all $\eta \in \R$. Fix $\bar{\eta}$ and choose $s_z^*$ such that $\frac{\partial \Ell}{\partial s_z}(\bar{\eta},s_z^*) = 0$. Then, by applying the implicit function theorem to $\frac{\partial \Ell}{\partial s_z}$, one obtains an open interval $(-\delta, \delta) \subseteq \R$ containing $\bar{\eta}$ and a unique function $\phi : (-\delta, \delta) \rightarrow \R^d$ such that $\phi(\bar{\eta}) = s_z^*$ and such that for all $\eta \in (-\delta, \delta)$, $\phi(\eta)$ is the unique minimizer of $s_z \mapsto \Ell(\eta,s_z)$. Moreover, $\phi$ is $C^1$ with the following derivative:

    \begin{equation}\label{eq:mlo-lemma-formula}
        \frac{\partial}{\partial \eta} \phi(\eta)
        = -\left[\frac{\partial^2 \Ell}{\partial s_z^2} (\eta, \phi(\eta))  \right]^{-1} \frac{\partial \Ell_2}{\partial s_z} (\phi(\eta)).
    \end{equation}
\end{lemma}


To apply this lemma, we will first show how to specify a score function $s_t(z)$ of the form in \Eqref{eq:score-function} evaluated at any $z \in \R^d$ as the minimizer of a score-matching objective. For the sake of simplicity, we will assume a constant scale schedule $\alpha_t \equiv 1$; our argument can be easily adapted to arbitrary scale schedules at the cost of additional notation. 

Let $s_t(z) : \R^d \rightarrow \R^d$ be the score function for some distribution $\rho_t := \rho * \mathcal{N}(0,\sigma_t^2 I)$, where $\rho$ is a target distribution on $\R^d$. \citet[Eqs. 14, 19]{kadkhodaie2024generalization} and the variational characterization of conditional expectation imply that this score function has the following pointwise variational characterization:

\begin{align*}
    s_t(z) &= \nabla \log \rho_t(z) \\
    &= \int_{\R^d} \left(\frac{x-z}{\sigma_t^2} \right) p(x|z)\textrm{d}x \\
    &= \underset{s(z)}{\textrm{argmin}} \int \frac 1 2 \left\|\frac{x-z}{\sigma_t^2} - s(z) \right\|^2 p(x|z)\textrm{d}x,
\end{align*}

where $p(x|z)$ is the conditional distribution of $x$ given $z \sim \rho_t$. While $p(x|z)$ is intractable a priori, $p(z|x) \sim \mathcal{N}(x,\sigma_t^2 I)$ is Gaussian, so we rewrite $p(x|z)$ in this integral using Bayes' theorem:

\begin{align*}
    s_t(z) &= \underset{s_z \in \R^d}{\textrm{argmin}} \int \frac 1 2 \left\|\frac{x-z}{\sigma_t^2} - s_z \right\|^2 p(x|z)\textrm{d}x \\
    &= \underset{s_z \in \R^d}{\textrm{argmin}} \int \frac 1 2 \left\|\frac{x-z}{\sigma_t^2} - s_z \right\|^2 \frac{\mathcal{N}(z;x,\sigma_t^2 I)}{\rho_t(z)}\rho(x)\textrm{d}x \\
    &= \underset{s_z \in \R^d}{\textrm{argmin}}
    \underset{x \sim \rho}{\E} \left[\frac{\mathcal{N}(z;x,\sigma_t^2 I)}{\rho_t(z)} \frac 1 2 \left\|\frac{x-z}{\sigma_t^2} - s_z \right\|^2  \right] \tag{$SM$}.
\end{align*}

Here, we use $\mathcal{N}(z;x,\sigma_t^2 I)$ to denote the density of a Gaussian distribution with mean $x$ and covariance $\sigma_t^2I$ evaluated at $z \in \R^d$. This provides a pointwise definition of the score $s_t(z)$ of $\rho_t$ evaluated at $z \in \R^d$ as the minimizer of the score-matching problem $(SM)$. In particular, applying this argument to the target distribution $\rho^\eta$ shows that:

\begin{equation*}
    s^\eta_t(z) = \underset{s_z \in \R^d}{\textrm{argmin}}
    \underset{x \sim \rho^\eta}{\E} \left[\frac{\mathcal{N}(z;x,\sigma_t^2 I)}{\rho^\eta_t(z)} \frac 1 2 \|\frac{x-z}{\sigma_t^2} - s_z \|^2  \right].
\end{equation*}

Now, define the following objective functions, in which we take $z \in \R^d$ to be fixed:

\begin{equation*}
    \Ell_\rho(s_z) = 
    \underset{x \sim \rho}{\E} \left[\frac{\mathcal{N}(z;x,\sigma_t^2 I)}{\rho^\eta_t(z)} \frac 1 2 \left\|\frac{x-z}{\sigma_t^2} - s_z \right\|^2  \right]
\end{equation*}

and

\begin{equation*}
    \Ell_\nu(s_z) = 
    \underset{x \sim \nu}{\E} \left[\frac{\mathcal{N}(z;x,\sigma_t^2 I)}{\rho^\eta_t(z)} \frac 1 2 \left\|\frac{x-z}{\sigma_t^2} - s_z \right\|^2  \right].
\end{equation*}

Using these two objectives, we can define an objective $\Ell(\eta,s_z) := \underbrace{\Ell_\rho(s_z)}_{:=\Ell_1} + \eta \underbrace{(\Ell_\nu(s_z) - \Ell_\rho(s_z))}_{:=\Ell_2}$ whose minimizer is $s^\eta_t(z)$. This objective is in the additive form prescribed by Lemma \ref{thm:mlo-lemma}, which will put us in position to apply the lemma once we verify that its remaining hypotheses are satisfied.

To this end, note that $\Ell$ is clearly a $C^2$ function of the prescribed additive form. 
Furthermore, as we will see below via a Hessian computation, the map $s_z \mapsto \Ell(\eta, s_z)$ is strictly convex for all $\eta \in \R$ and for all $z \in \R^d$. 
Fixing a point $(\bar{\eta}, s_z^*) = (\bar{\eta}, s^\eta_t(z))$ yields a critical point of $\Ell$ with respect to $s_z$, which puts us in position to apply Lemma \ref{thm:mlo-lemma}.

Lemma \ref{thm:mlo-lemma} gives us a function $\phi(\eta)$ defined on an open set containing $\bar{\eta}$ that maps $\eta$ to the unique minimizer $s^\eta_t(z)$ of $\Ell(\eta, s_z)$. Crucially, it gives us a formula for the derivative $\frac{\partial}{\partial \eta} \phi(\eta)$, which involves the derivative $\frac{\partial \Ell_2}{\partial s_z} (\phi(\eta))$ and the Hessian $\frac{\partial^2 \Ell}{\partial s_z^2} (\eta, \phi(\eta))$. We will compute each of these terms separately.

We begin by computing the derivative $\frac{\partial \Ell_2}{\partial s_z} (\phi(\eta))$. We have

\begin{align*}
    \frac{\partial \Ell_2}{\partial s_z} (\phi(\eta)) &= \frac{\partial}{\partial s_z} \left[ \underset{x \sim \nu}{\E} \left[\frac{\mathcal{N}(z;x,\sigma_t^2 I)}{\rho^\eta_t(z)} \frac 1 2 \left\|\frac{x-z}{\sigma_t^2} - s_z \right\|^2  \right] - \underset{x \sim \rho}{\E} \left[\frac{\mathcal{N}(z;x,\sigma_t^2 I)}{\rho^\eta_t(z)} \frac 1 2 \left\|\frac{x-z}{\sigma_t^2} - s_z \right\|^2  \right] \right]\Bigg|_{s^\eta_t(z)} \\
    &= \frac{\partial}{\partial s_z} \underset{x \sim (\nu - \rho)}{\E} \left[\frac{\mathcal{N}(z;x,\sigma_t^2 I)}{\rho^\eta_t(z)} \frac 1 2 \left\|\frac{x-z}{\sigma_t^2} - s_z \right\|^2  \right]\Bigg|_{s^\eta_t(z)} \\
    &= \underset{x \sim (\nu - \rho)}{\E} \left[\frac{\mathcal{N}(z;x,\sigma_t^2 I)}{\rho^\eta_t(z)} \frac{\partial}{\partial s_z} \frac 1 2 \left\|\frac{x-z}{\sigma_t^2} - s_z \right\|^2  \right]\Bigg|_{s^\eta_t(z)} \\
    &= \underset{x \sim \nu}{\E} \left[-\frac{\mathcal{N}(z;x,\sigma_t^2 I)}{\rho^\eta_t(z)} \left( \frac{x-z}{\sigma_t^2} - s_z \right)  \right] \Bigg|_{s^\eta_t(z)} - \underset{x \sim \rho}{\E} \left[-\frac{\mathcal{N}(z;x,\sigma_t^2 I)}{\rho^\eta_t(z)} \left( \frac{x-z}{\sigma_t^2} - s_z \right)  \right] \Bigg|_{s^\eta_t(z)} \\
    &= \underset{x \sim \rho}{\E} \left[\frac{\mathcal{N}(z;x,\sigma_t^2 I)}{\rho^\eta_t(z)} \left( \frac{x-z}{\sigma_t^2} - s^\eta_t(z) \right)  \right] - \underset{x \sim \nu}{\E} \left[\frac{\mathcal{N}(z;x,\sigma_t^2 I)}{\rho^\eta_t(z)} \left( \frac{x-z}{\sigma_t^2} - s^\eta_t(z) \right)  \right]
\end{align*}

We now rewrite each expectation in the last line in terms of the scores of $\rho_t$ and $\nu_t$. To rewrite the first expectation, we pull out the factor of $\frac{1}{\rho^\eta_t(z)}$, which does not depend on $x$, and multiply by $1 = \frac{\rho_t(z)}{\rho_t(z)}$ to obtain the following:

\begin{align*}
    \underset{x \sim \rho}{\E} \left[\frac{\mathcal{N}(z;x,\sigma_t^2 I)}{\rho^\eta_t(z)} \left( \frac{x-z}{\sigma_t^2} - s^\eta_t(z) \right)  \right] 
    &= \frac{\rho_t(z)}{\rho^\eta_t(z)} \underset{x \sim \rho}{\E} \left[\frac{\mathcal{N}(z;x,\sigma_t^2 I)}{\rho_t(z)} \left( \frac{x-z}{\sigma_t^2} - s^\eta_t(z) \right)  \right] \\
    &= \frac{\rho_t(z)}{\rho^\eta_t(z)} \left(
    \underbrace{\underset{x \sim \rho}{\E} \left[\frac{\mathcal{N}(z;x,\sigma_t^2 I)}{\rho_t(z)} \left( \frac{x-z}{\sigma_t^2} \right)\right]}_{=s^\rho_t(z)}
    - \underbrace{\underset{x \sim \rho}{\E} \left[\frac{\mathcal{N}(z;x,\sigma_t^2 I)}{\rho_t(z)}\right]}_{=1} s^\eta_t(z) 
    \right) \\
    &= \frac{\rho_t(z)}{\rho^\eta_t(z)}\left(s^\rho_t(z) - s^\eta_t(z) \right).
\end{align*}

Analogous reasoning allows us to conclude that

\begin{equation*}
    \underset{x \sim \nu}{\E} \left[\frac{\mathcal{N}(z;x,\sigma_t^2 I)}{\rho^\eta_t(z)} \left( \frac{x-z}{\sigma_t^2} - s^\eta_t(z) \right)  \right]
    = \frac{\nu_t(z)}{\rho^\eta_t(z)}\left(s^\nu_t(z) - s^\eta_t(z) \right),
\end{equation*}

and putting these together, we obtain

\begin{equation}\label{eq:dL2/dsz}
    \frac{\partial \Ell_2}{\partial s_z} (\phi(\eta))
    = \frac{\rho_t(z)}{\rho^\eta_t(z)} s^\rho_t(z) - \frac{\nu_t(z)}{\rho^\eta_t(z)} s^\nu_t(z) 
    + \left(\frac{\nu_t(z) - \rho_t(z)}{\rho^\eta_t(z)} \right)s^\eta_t(z).
\end{equation}

We now compute the Hessian term $\frac{\partial^2 \Ell}{\partial s_z^2} (\eta, \phi(\eta))$. Note that $\frac{\partial^2 \Ell}{\partial s_z^2} = \frac{\partial^2}{\partial s_z^2} \Ell_\rho + \eta (\frac{\partial^2}{\partial s_z^2} \Ell_\nu - \frac{\partial^2}{\partial s_z^2} \Ell_\rho)$, and that we have already computed the relevant first derivatives:

\begin{equation*}
    \frac{\partial}{\partial s_z} \Ell_\rho
    = \underset{x \sim \rho}{\E} \left[-\frac{\mathcal{N}(z;x,\sigma_t^2 I)}{\rho^\eta_t(z)} \left( \frac{x-z}{\sigma_t^2} - s_z \right)  \right]
\end{equation*}

and 

\begin{equation*}
    \frac{\partial}{\partial s_z} \Ell_\nu
    = \underset{x \sim \nu}{\E} \left[-\frac{\mathcal{N}(z;x,\sigma_t^2 I)}{\rho^\eta_t(z)} \left( \frac{x-z}{\sigma_t^2} - s_z \right)  \right].
\end{equation*}

Differentiating again and simplifying, we see that

\begin{equation*}
    \frac{\partial^2}{\partial s_z^2} \Ell_\rho = \frac{\rho_t(z)}{\rho^\eta_t(z)}I
\end{equation*}

and

\begin{equation*}
    \frac{\partial^2}{\partial s_z^2} \Ell_\nu = \frac{\nu_t(z)}{\rho^\eta_t(z)}I.
\end{equation*}

Combining these and noting that $\rho^\eta_t = (1-\eta)\rho_t + \eta \nu_t$, we conclude that $\frac{\partial^2 \Ell}{\partial s_z^2} (\eta, s_z) = I$ for all $\eta \in \R$ and for all $s_z$. In particular, the map $s_z \mapsto \Ell(\eta,s_z)$ is strictly convex for all $\eta \in [0,1]$ as required by Lemma \ref{thm:mlo-lemma}.

We finally substitute these first and second derivatives into \Eqref{eq:mlo-lemma-formula} to obtain:

\begin{align*}
    \frac{\partial}{\partial \eta} \phi(\eta)
        &= -\left[\frac{\partial^2 \Ell}{\partial s_z^2} (\eta, \phi(\eta))  \right]^{-1} \frac{\partial \Ell_2}{\partial s_z} (\phi(\eta)) \\
        &= -[I]^{-1} \left( \frac{\rho_t(z)}{\rho^\eta_t(z)} s^\rho_t(z) - \frac{\nu_t(z)}{\rho^\eta_t(z)} s^\nu_t(z) 
    + \left(\frac{\nu_t(z) - \rho_t(z)}{\rho^\eta_t(z)} \right)s^\eta_t(z) \right) \\
    &= \frac{\nu_t(z)}{\rho^\eta_t(z)} s^\nu_t(z) - \frac{\rho_t(z)}{\rho^\eta_t(z)} s^\rho_t(z) 
    + \left(\frac{\rho_t(z) - \nu_t(z)}{\rho^\eta_t(z)} \right)s^\eta_t(z).
\end{align*}

In particular, if $\eta = 0$, then $\rho^\eta_t(z) = \rho_t(z)$ and this simplifies to:

\begin{equation*}
     \frac{\partial}{\partial \eta} \phi(\eta) = \frac{\nu_t(z)}{\rho_t(z)}\left(s^\nu_t(z) - s^\rho_t(z) \right) =: g_t(z).
\end{equation*}

This completes the first part of the proof.

\subsubsection{$g_t$ is the Fr\'{e}chet derivative of $T_t(\eta) : \eta \mapsto s^\eta_t$ at $\eta=0$}

We now extend this pointwise argument to the space of functions. Consider the map $T_t : \R \rightarrow L^2(\R^d, \rho^\eta_t)$ that maps $\eta$ to $s^\eta_t$. We will show that $g_t$ is the Fr\'{e}chet derivative of $T_t$ at $\eta=0$ for any $t \in [t_0, t_1]$. To do so, we need to show that for any $t \in [t_0,t_1]$,

\begin{equation*}
    \lim_{h \rightarrow 0} \left\|\frac{s^{h}_t - s^{0}_t}{h} - g_t \right\|_{L^2(\R^d, \rho^0_t)} = 0.
\end{equation*}

The previous section shows that $g_t(z)$ is the \emph{pointwise} derivative of $s^\eta_t$ with respect to $\eta$ at $\eta=0$. This means that for any $z \in \R^d, t\in[t_0,t_1]$,

\begin{equation*}
    \frac{s^{h}_t(z) - s^{0}_t(z)}{h} \rightarrow g_t(z).
\end{equation*}

Hence $\frac{s^{h}_t - s^{0}_t}{h}$ converges pointwise to $g_t$ for all $t \in [t_0,t_1]$. We will use the dominated convergence theorem (DCT) to lift this pointwise convergence to $L^2(\R^d, \rho^0_t)$ convergence. Define the following function:

\begin{equation*}
    F_h(z;t) := \frac{s^{h}_t(z) - s^{0}_t(z)}{h}
\end{equation*}

We need to show that there exists some real-valued function $G(z;t) \in L^2(\R^d, \rho^0_t)$ such that $\|F_h(z;t)\|_2 \leq G(z;t)$ uniformly in $h$ for all $z,t$. To this end, note that by the mean value theorem, there exists some $\theta \in [0,1]$ such that:

\begin{align*}
    \|F_h(z;t)\|_2 &= \left\|\frac{s^{h}_t(z) - s^{0}_t(z)}{h} \right\|_2 \\
    &\leq \left\|\frac{\partial}{\partial\eta} s^\eta_t(z) \Big|_{\eta = \theta h} \right\|_2 \\
    &= \left\|\frac{\rho_t(z)}{\rho^{\theta h}_t(z)}\left(s^\rho_t(z) - s^{\theta h}_t(z) \right) 
    - \frac{\nu_t(z)}{\rho^{\theta h}_t(z)}\left(s^\nu_t(z) - s^{\theta h}_t(z) \right) \right\|_2,
\end{align*}

where the last line follows from a rearrangement of \Eqref{eq:dL2/dsz}. We can further simplify this bound to eliminate the dependence on $h$. First, note that $\rho^{\theta h}_t = (1-\theta h)\rho_t + \theta h \nu_t$, so that for $h \leq \frac 1 2$, we have

\begin{equation*}
    \frac{1}{\rho^{\theta h}_t(z)} = 
    \frac{1}{(1-\theta h)\rho_t(z) + \theta h \nu_t(z)}
    \leq \frac{1}{(1-\theta h)\rho_t(z)}
    \leq \frac{2}{\rho_t(z)}.
\end{equation*}

Hence, for $h$ sufficiently small, we have:

\begin{multline*}
    \|F_h(z;t)\|_2 \leq \left\|\frac{\rho_t(z)}{\rho^{\theta h}_t(z)}\left(s^\rho_t(z) - s^{\theta h}_t(z) \right) 
    - \frac{\nu_t(z)}{\rho^{\theta h}_t(z)}\left(s^\nu_t(z) - s^{\theta h}_t(z) \right) \right\|_2 \\
    \leq 
    \frac{2}{\rho_t(z)}\left\|\rho_t(z)\left(s^\rho_t(z) - s^{\theta h}_t(z) \right) 
    - \nu_t(z)\left(s^\nu_t(z) - s^{\theta h}_t(z) \right) \right\|_2.
\end{multline*}

Applying the triangle inequality, we then obtain:

\begin{multline*}
    \frac{2}{\rho_t(z)}\left\|\rho_t(z)\left(s^\rho_t(z) - s^{\theta h}_t(z) \right) 
    - \nu_t(z)\left(s^\nu_t(z) - s^{\theta h}_t(z) \right) \right\|_2 \\
    \leq 
    \frac{2}{\rho_t(z)} \left(\rho_t(z) \| s^\rho_t(z) - s^{\theta h}_t(z) \|_2 
    + \nu_t(z) \| s^\nu_t(z) - s^{\theta h}_t(z) \|_2 \right).
\end{multline*}

Now, define

\begin{equation*}
    k^\rho_t(z) := \int w_t(z,x)x\textrm{d}\rho(x)
\end{equation*}

and similarly for $k^\nu_t(z)$ and $k^{\theta h}_t(z)$. Then \Eqref{eq:score-function} tells us that

\begin{equation*}
    s^\rho_t(z) = \frac{1}{\sigma_t^2}\left(k^\rho_t(z) - z \right),
\end{equation*}

and similar identities hold for the other score functions. Furthermore,

\begin{equation*}
    \| s^\rho_t(z) - s^{\theta h}_t(z) \|_2
    = \frac{1}{\sigma_t^2}\| k^\rho_t(z) - k^{\theta h}_t(z) \|_2
    \leq \frac{1}{\sigma_t^2} \left(\| k^\rho_t(z) \|_2 + \|k^{\theta h}_t(z) \|_2 \right),
\end{equation*}

where the last line follows from the triangle inequality. Because $k^\rho_t(z)$ is a convex combination of points in the compact support of $\rho$, we can bound $\| k^\rho_t(z) \|_2 \leq D^\rho < +\infty$, where $D^\rho$ is the diameter of the support of $\rho$. Similarly, $\| k^\nu_t(z) \|_2 \leq D^\nu < +\infty$, and because $\textrm{supp}(\rho^{\theta h}) \subseteq \textrm{supp}(\rho)\cup\textrm{supp}(\nu)$, we have $\|k^{\theta h}_t(z) \|_2 \leq D^\rho + D^\nu$. Substituting these bounds into the above and simplifying, we obtain:

\begin{align*}
    \|F_h(z;t)\|_2 &\leq 
    \frac{2}{\rho_t(z)} \left(\rho_t(z) \| s^\rho_t(z) - s^{\theta h}_t(z) \|_2 
    + \nu_t(z) \| s^\nu_t(z) - s^{\theta h}_t(z) \|_2 \right) \\
    &\leq \frac{2}{\rho_t(z)} \left(\frac{\rho_t(z)}{\sigma_t^2}(2D^\rho + D^\nu) 
    + \frac{\nu_t(z)}{\sigma_t^2}(2D^\nu + D^\rho)  \right) \\
    &=: G(z;t)
\end{align*}

This function $G(z;t)$ dominates $\|F_h(z;t)\|_2$ uniformly in $h$ for all $z,t$. It remains to show that $G(z;t) \in L^2(\R^d, \rho^0_t)$. To this end, first note that $\rho^0_t = \rho_t$. Then,

\begin{align*}
    \int G(z;t)\textrm{d}\rho^0_t(z) &= \int G(z;t)\textrm{d}\rho_t(z) \\
    &= \int \frac{2}{\rho_t(z)} \left(\frac{\rho_t(z)}{\sigma_t^2}(2D^\rho + D^\nu) 
    + \frac{\nu_t(z)}{\sigma_t^2}(2D^\nu + D^\rho)  \right) \rho_t(z)\textrm{d}z \\
    &= \frac{2(2D^\rho + D^\nu)}{\sigma_t^2}
    \underbrace{\int \rho_t(z)\textrm{d}z}_{=1} + \frac{2(2D^\nu + D^\rho)}{\sigma_t^2}\underbrace{\int \nu_t(z)\textrm{d}z}_{=1} \\
    &= \frac{6}{\sigma_t^2}(D^\rho + D^\nu) \\
    &< +\infty.
\end{align*}

This shows that $G(z;t) \in L^2(\R^d, \rho^0_t)$. As the hypotheses of the DCT are satisfied, we finally conclude that $g_t$ is the Fr\'{e}chet derivative of $T_t$ at $\eta=0$ for any $t \in [t_0, t_1]$. This completes the proof of Theorem \ref{thm:score-sensitivity}. $\blacksquare$

\section{Visualizing the sample sensitivities}\label{sec:sample-sensitivity-viz}

\begin{figure*}[h]
    \centering
    \includegraphics[scale=0.25]{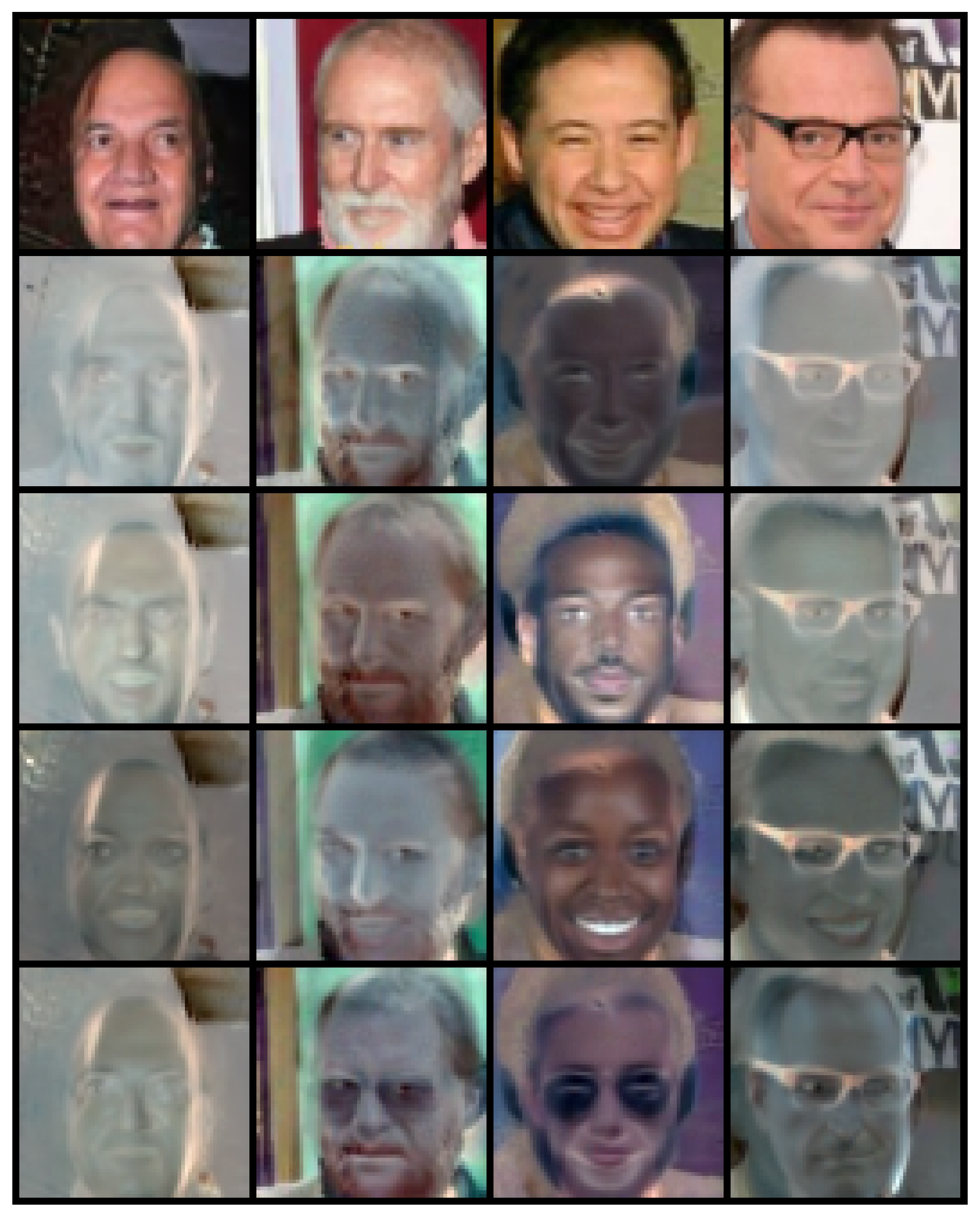}
    \caption{The bottom four rows depict solutions $\frac{\textrm d}{\textrm d \eta} \Phi_{\tilde{t}_1}^\eta(z_0)\Big|_{\eta = 0}$ to the sample sensitivity equation \plaineqref{eq:ode-sensitivity} for model samples $\Phi_{\tilde{t}_1}^0(z_0)$ pictured in the top row. In each of the lower four rows, the perturbation measure $\nu$ is the empirical distribution over images from the CelebA test set with attributes ``bald'', ``goatee'', ``smiling'', and ``eyeglasses'', respectively.}
    \label{fig:sample-sensitivity-four-attr}
\end{figure*}

In this appendix, we illustrate our sample sensitivity analysis on images from the CelebA dataset. We draw four samples from a base model trained on CelebA and solve \Eqref{eq:ode-sensitivity} for each model sample and for four perturbation measures $\nu$, each of which is an empirical measure over a perturbation set $S$. These perturbation sets consist of samples from the CelebA test set possessing the attribute labels ``bald'', ``goatee'', ``smiling'', and ``eyeglasses'', respectively. We depict model samples in the top row of Figure \ref{fig:sample-sensitivity-four-attr} and solutions to the sample sensitivity ODE for each perturbation set in the bottom four rows. Solutions to this ODE should approximate changes in the model samples in the top row in response to perturbing the base model's target distribution, and many of these predictions are intuitively reasonable in practice. For instance, base model samples representing people without glasses are pushed towards samples of people with glasses in response to perturbing the target distribution towards CelebA samples with the ``eyeglasses'' attribute, and one observes similar phenomena for the other perturbation sets.

\begin{figure*}[h]
    \centering
    \subfigure[``Bald'']{\includegraphics[scale=0.35]{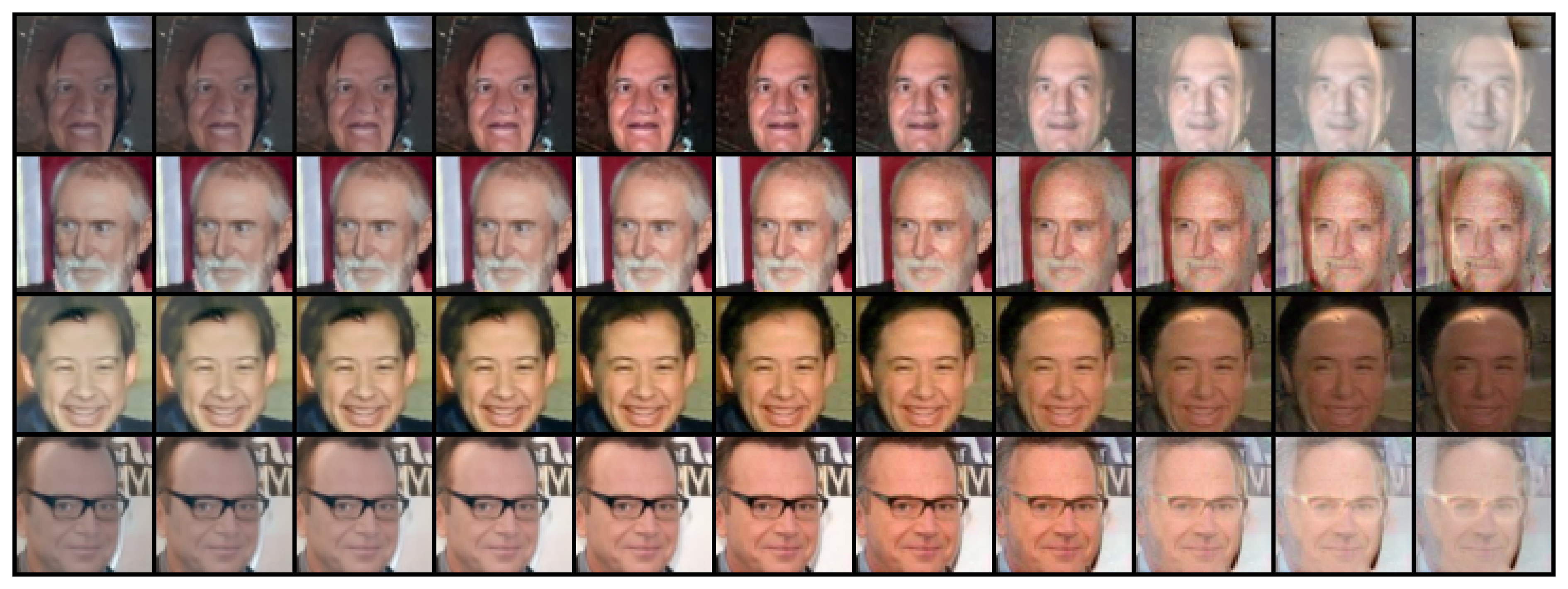}}
    \subfigure[``Goatee'']{\includegraphics[scale=0.35]{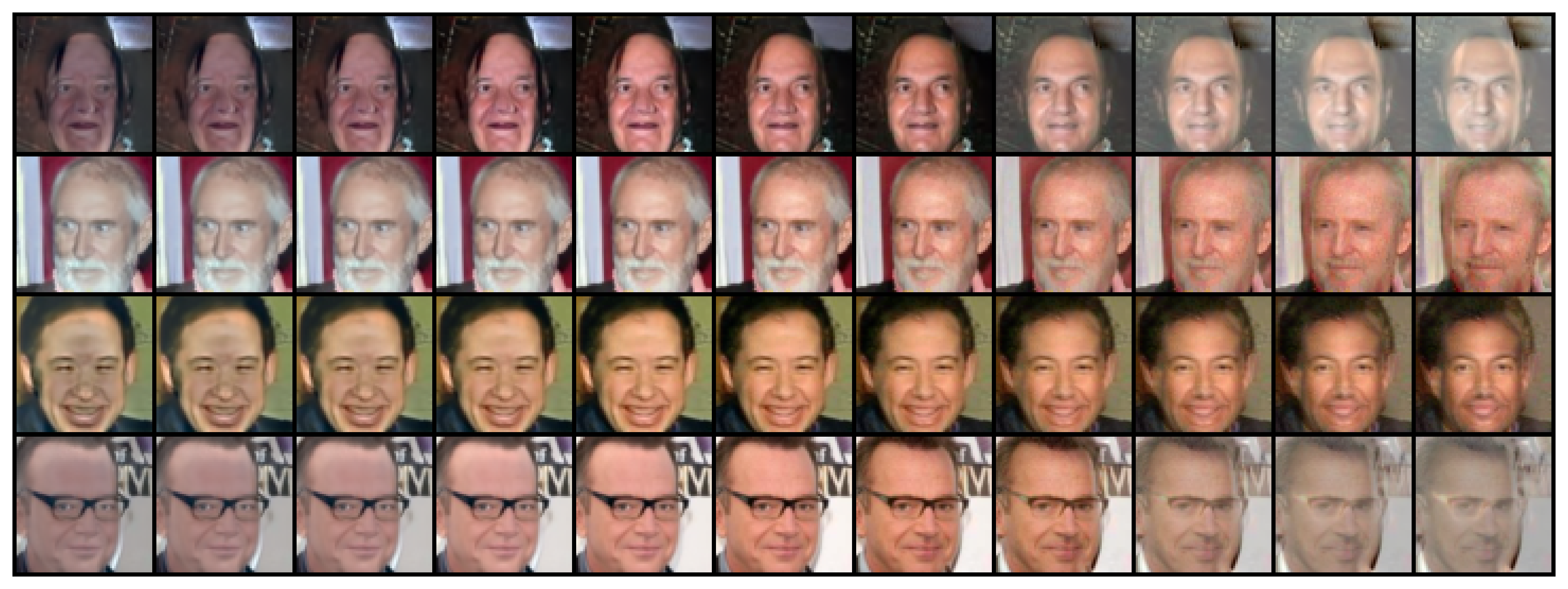}}
    \subfigure[``Smiling'']{\includegraphics[scale=0.35]{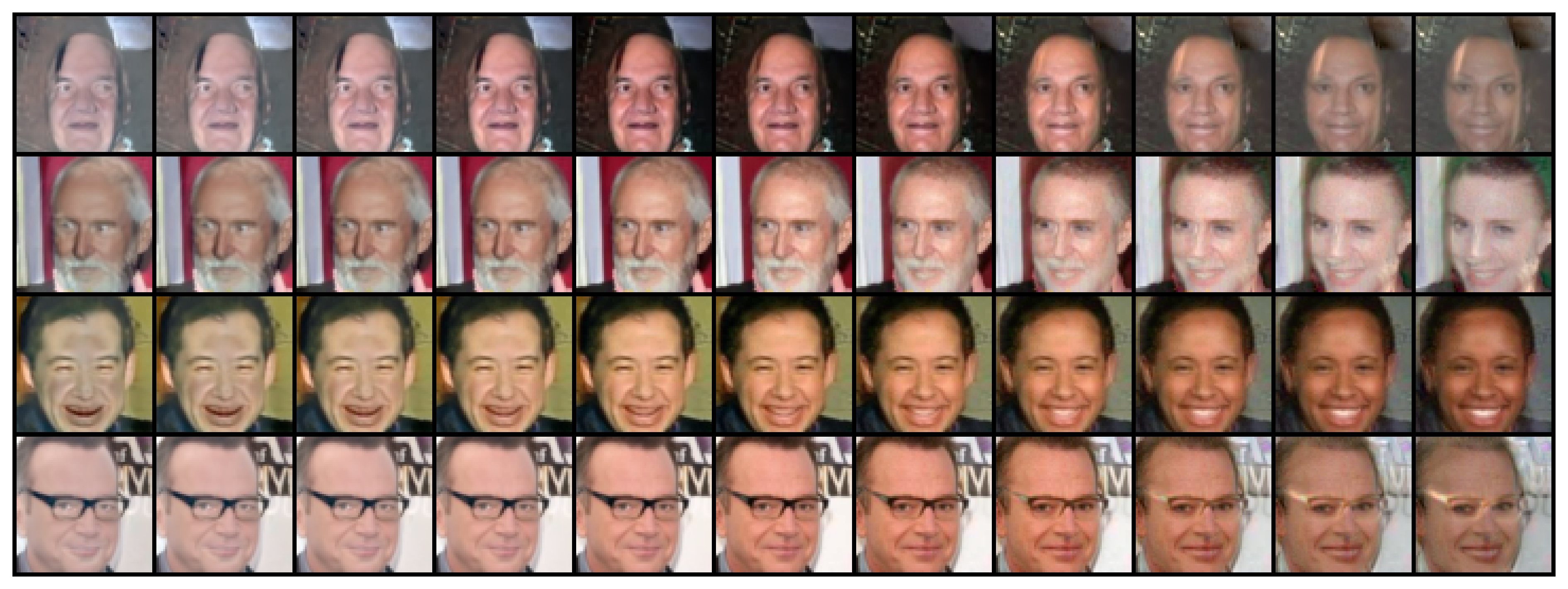}}
    \subfigure[``Eyeglasses'']{\includegraphics[scale=0.35]{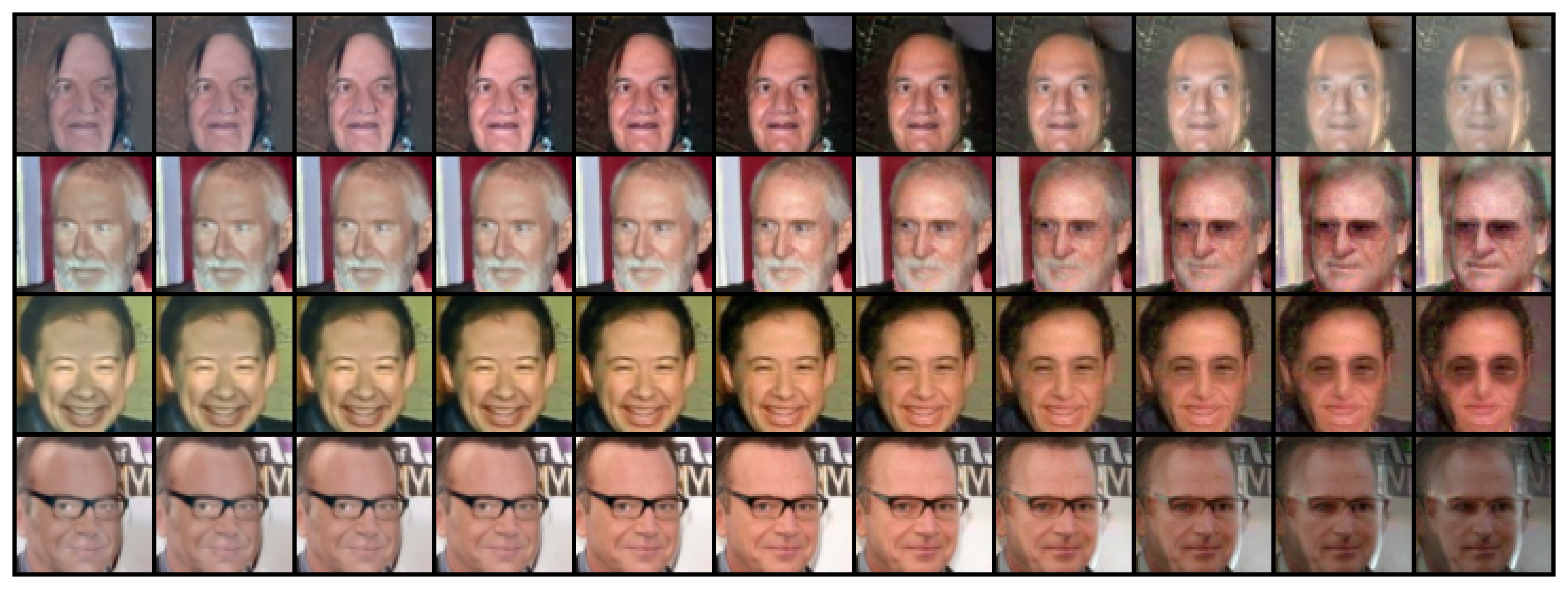}}
    \caption{Line segments extending from model samples (center images) towards negative (left) and positive (right) multiples of sample sensitivities $\frac{\textrm d}{\textrm d \eta} \Phi_{\tilde{t}_1}^\eta(z_0)\Big|_{\eta = 0}$. In each subfigure, the perturbation measure $\nu$ is the empirical distribution over CelebA test samples with the attribute listed in the subcaption.}
    \label{fig:sample-sensitivity-rays}
\end{figure*}

In Figure \ref{fig:sample-sensitivity-rays}, we also depict line segments of the form $\Phi_{\tilde{t}_1}^0(z_0) + \alpha \frac{\textrm d}{\textrm d \eta} \Phi_{\tilde{t}_1}^\eta(z_0)\Big|_{\eta = 0}$ for $\alpha \in [-2,2]$ and for the sample sensitivity ODE solutions depicted in Figure \ref{fig:sample-sensitivity-four-attr}. These line segments should approximate samples from a model whose target distributed has been perturbed towards $\pm \nu$, where $\nu$ is the empirical measure over CelebA test images with the specified attributes. For $\alpha$ close to 0, the perturbed samples resemble the original sample (6th from the left in each row), differing mainly in the strength of the specified attribute. As $\alpha$ moves farther from 0, the perturbed samples deviate increasingly from the original.

\section{Experiment details}\label{sec:experiment-details}

\subsection{Synthetic experiments}\label{sec:synthetic-exper-details}

\subsubsection{First-order approximation for perturbed model samples}\label{sec:first-order-exper-details}

In this experiment, the initial target measure $\rho$ is an equally-weighted mixture of two Gaussians on $\R^{100}$ with means $(-1,...,-1)$ and $(1,...,1)$, respectively,  and shared covariance $\sigma^2 I$ for $\sigma = 0.1$. We perturb $\rho$ in the direction of a Gaussian distribution $\nu$ centred at $(1,...,1)$ with covariance $\sigma^2 I$ for $\sigma=0.1$. For any $\bar{\eta} \in [0,1]$, the perturbed target $\rho^{\bar{\eta}} = (1-\bar{\eta})\rho +\bar{\eta} \nu$ is a mixture of Gaussians with the same means and covariances as $\rho$, but with weights $\frac{1-\bar{\eta}}{2}$ and $\frac{1+\bar{\eta}}{2}$.

We obtain sample paths $z_t$ for $\rho_t$ and $\rho^{\bar{\eta}}_t$ by fixing 1000 base samples $z_0 \sim \rho_0$ and numerically integrating the PF-ODE and variance-preserving SDE using a forward Euler scheme and Euler-Maruyama scheme, resp., with step sizes $\Delta t \in \{1\times10^{-4}, 5\times10^{-4}, 1\times10^{-3}, 5\times10^{-3} \}$. Our scale and noise schedules come from a linear \texttt{DDPMScheduler} from the \texttt{diffusers} library with $\beta_{\textrm{start}} = 10^{-4}$ and $\beta_{\textrm{end}} = 0.02$. We exactly compute the Gaussian mixture densities $\rho_t(z_t)$ along each sample path. We then integrate \Eqref{eq:ode-sensitivity} using the same forward Euler scheme to obtain the sensitivities $\frac{\textrm d}{\textrm d \eta} \Phi_{\tilde{t}_1}^\eta(z_0)\Big|_{\eta = 0}$ of samples from $\rho_{\tilde{t}_1}$. We compute the Taylor remainder:

\begin{equation*}
    R(\bar{\eta}) := \left(\Phi_{\tilde{t}_1}^{\bar{\eta}}(z_0) -\Phi_{\tilde{t}_1}^0(z_0) \right) -  \bar{\eta} \frac{\textrm d}{\textrm d \eta} \Phi_{\tilde{t}_1}^\eta(z_0)\Big|_{\eta = 0}
\end{equation*}

and report the median value of $\frac{R(\bar{\eta})}{\bar{\eta}}$ across the 1000 batch samples in our plots.

In our experiments studying the effect of using Hutchinson's estimator to estimate model densities, we use the same setup as in the step size experiments, but estimate the base model densities $\rho_t(z)$ with Hutchinson's estimator: $\textrm{tr}(A) = \E[\epsilon^\top A \epsilon]$. We use standard normal Gaussian samples for $\epsilon$ and report the number of noise samples we used in our plots.

\subsubsection{Stability under score approximation error}\label{sec:stability-exper-details}

Here, the initial target measure $\rho$ is an equally-weighted mixture of two Gaussians on $\R^{10}$ with means $(-1,...,-1)$ and $(1,...,1)$  and shared covariance $\sigma^2 I$ for $\sigma = 0.1$. We perturb $\rho$ in the direction of a Gaussian distribution $\nu$ centred at $(1,...,1)$ with covariance $\sigma^2 I$ for $\sigma=0.1$. Instead of evaluating the score of $\rho_t$ in closed form as in \ref{sec:stepsize-hutch-val}, we now train a neural network to approximate this score function. Our neural network is a two-hidden-layer MLP with SiLU activations and 512-dimensional hidden layers. We also use Fourier features \citep{tancik2020fourfeat} with 128 frequencies and $\sigma=2.0$. We solve the score-matching problem using AdamW with a learning rate of $10^{-4}$ and a batch size of 100k. We train for 200k steps in total. In our plot, we omit the first two measurements of the correlations for clarity, as the training loss was large and network was very far from convergence during this phase of training .

We fix 1000 base samples $z_0 \sim \rho_0$ and evaluate the sensitivity of model samples from the exact diffusion model $\rho_t$ and its neural approximation every 1000 training steps. We discretize all ODEs using a forward Euler scheme with step size $10^{-2}$ and use Hutchinson's estimator with 100 samples to estimate the model densities $\rho_t(z)$. We measure the median correlation between the exact and approximate sample sensitivities and compare it to the value of the score-matching loss at that training step in Figure \ref{fig:sample-sensitivity-correlations}.

\subsection{Image datasets}\label{sec:image-exper-details}

\paragraph{Retraining experiments.}
Each neural diffusion model in these experiments is parametrized by a \texttt{Unet2DModel} from the \texttt{diffusers} library. For the CelebA experiments, we set \texttt{layers\_per\_block=2}, \texttt{block\_out\_channels=(128, 256, 512, 512)}, and \texttt{norm\_num\_groups=32}. We use a \texttt{DDPMScheduler} with $\beta_{\textrm{start}} = 10^{-4}$ and $\beta_{\textrm{end}} = 0.02$. The base model samples consist of 10k iid samples from the CelebA training set, and the new samples $S$ are 495 CelebA training samples with a large CLIP score for ``a photo of an old man''. We pre-process the training images by center-cropping to a size of $140\times140$, then resizing to $64\times64$ and normalizing to $[-1,1]$. We apply random horizontal flips as augmentations in training. We then train the CelebA diffusion models for 1000 epochs with an effective batch size of 512. Our optimizer is AdamW with a learning rate of $10^{-4}$.

For the MNIST experiments, we set \texttt{layers\_per\_block=2}, \texttt{block\_out\_channels=(32, 64, 128)}, and \texttt{norm\_num\_groups=8}. We use a \texttt{DDPMScheduler} with $\beta_{\textrm{start}} = 10^{-4}$ and $\beta_{\textrm{end}} = 0.02$. We do not apply any preprocessing to these samples. We train the MNIST diffusion models for 100 epochs with an effective batch size of 1024. Our optimizer is AdamW with a learning rate of $10^{-4}$.

We draw model samples by integrating the PF-ODE and estimate model densities along the sample path using Hutchinson's estimator with 1 sample. We numerically integrate the PF-ODE and our sample sensitivity ODE \plaineqref{eq:ode-sensitivity} using a forward Euler scheme with a step size of $10^{-3}$. We clamp the $\frac{\nu_t(z)}{\rho_t(z)}$ weights to $[0.1,10]$ for numerical stability. For the entropic OT baseline, we use the \texttt{sinkhorn\_log} algorithm from the \texttt{POT} package \citep{flamary2024pot} with a regularization value of $0.05$ to compute the coupling matrix.

\paragraph{Fine-tuning experiments}

These experiments mostly replicate the setup in our retraining experiments, but implement the following changes. For CelebA, we train the base model on 10k iid samples from the CelebA training set for 1k epochs with the same hyperparameters as in the retraining experiments, and then fine-tune for 200 epochs on 495 CelebA training samples with a large CLIP score for ``a photo of an old man''. We use the same learning rate of $10^{-4}$ for fine-tuning.

For MNIST, we train the base model on the MNIST training set for 100 epochs with an effective batch size of 1024 and a learning rate of $10^{-4}$, and then fine-tune on TMNIST for a single epoch at a learning rate of $10^{-5}$. 

\end{document}